\pdfoutput=1
\documentclass[11pt]{article}
\usepackage[table]{xcolor}

\usepackage[preprint]{acl}

\usepackage{times}
\usepackage{latexsym}
\usepackage{amsmath}
\usepackage{amssymb}
\usepackage{booktabs}
\usepackage{array}
\usepackage{comment}
\usepackage{multirow}
\usepackage[most]{tcolorbox} 
\usepackage{lipsum}  
\usepackage{stfloats}
\usepackage{diagbox}  % for diagonal cell in tables
\usepackage{algorithm}
\usepackage{algpseudocode}
\usepackage{subcaption}
\usepackage{arydshln}
\usepackage{dashrule}
\usepackage{colortbl}
\usepackage{url}

\newtcolorbox{blackbox}[1][]{
  enhanced,
  colback=white,
  colframe=black,
  coltitle=white,
  fonttitle=\bfseries,
  title=#1,
  boxrule=0.8pt,
  arc=4pt,
  top=1pt,
  bottom=1pt,
  left=6pt,
  right=6pt,
  width=\textwidth,
  colbacktitle=black,
  toptitle=1pt,
  bottomtitle=1pt
}

\definecolor{LightPink}{RGB}{255,182,193}
\definecolor{LightRed}{RGB}{255,204,203}
\definecolor{LightGray}{RGB}{245,245,245}
\definecolor{SuperLightPink}{rgb}{1.0, 0.88, 0.9}
\definecolor{LightPink}{rgb}{1, 0.9, 0.9}
\definecolor{SoftRed}{rgb}{1, 0.7, 0.7}
\definecolor{SoftGreen}{rgb}{0.8, 1, 0.8}
\definecolor{SoftPink}{RGB}{255, 204, 213}
\definecolor{SoftBlue}{RGB}{204, 229, 255}
\definecolor{SuperSoftPink}{RGB}{255, 230, 236}
\definecolor{SuperSoftBlue}{RGB}{230, 243, 255}
\usepackage[T1]{fontenc}
\usepackage[utf8]{inputenc}

\usepackage{microtype}

\usepackage{inconsolata}

\usepackage{graphicx}

\title{RECIPE-TKG: From Sparse History to Structured Reasoning for LLM-based Temporal Knowledge Graph Completion}
\author{
Ömer Faruk Akgül$^{1}$\thanks{Equal contribution.} \quad
Feiyu Zhu$^{1}$\footnotemark[1] \quad
Yuxin Yang$^{1}$ \quad
Rajgopal Kannan$^{2}$ \quad
Viktor Prasanna$^{1}$ \\
$^{1}$University of Southern California \\
$^{2}$DEVCOM ARL Army Research Office \\
\texttt{\{akgul, feiyuzhu, yyang393, prasanna\}@usc.edu}, \texttt{rajgopal.kannan.civ@army.mil}
}

\begin{document}
\maketitle
\begin{abstract}
Temporal Knowledge Graphs (TKGs) represent dynamic facts as timestamped relations between entities. TKG completion involves forecasting missing or future links, requiring models to reason over time-evolving structure. While LLMs show promise for this task, existing approaches often overemphasize supervised fine-tuning and struggle particularly when historical evidence is limited or missing. We introduce RECIPE-TKG, a lightweight and data-efficient framework designed to improve accuracy and generalization in settings with sparse historical context. It combines (1) rule-based multi-hop retrieval for structurally diverse history, (2) contrastive fine-tuning of lightweight adapters to encode relational semantics, and (3) test-time semantic filtering to iteratively refine generations based on embedding similarity. Experiments on four TKG benchmarks show that RECIPE-TKG outperforms previous LLM-based approaches, achieving up to 30.6\% relative improvement in Hits@10. Moreover, our proposed framework produces more semantically coherent predictions, even for the samples with limited historical context.

\let\thefootnote\relax
\footnotetext{The code is available at \href{https://github.com/farukakgul/Recipe-TKG}{github/farukakgul/Recipe-TKG}.}

\end{abstract}

\section{Introduction}

\begin{figure} [ht]
    \centering
    \includegraphics[width=\linewidth]{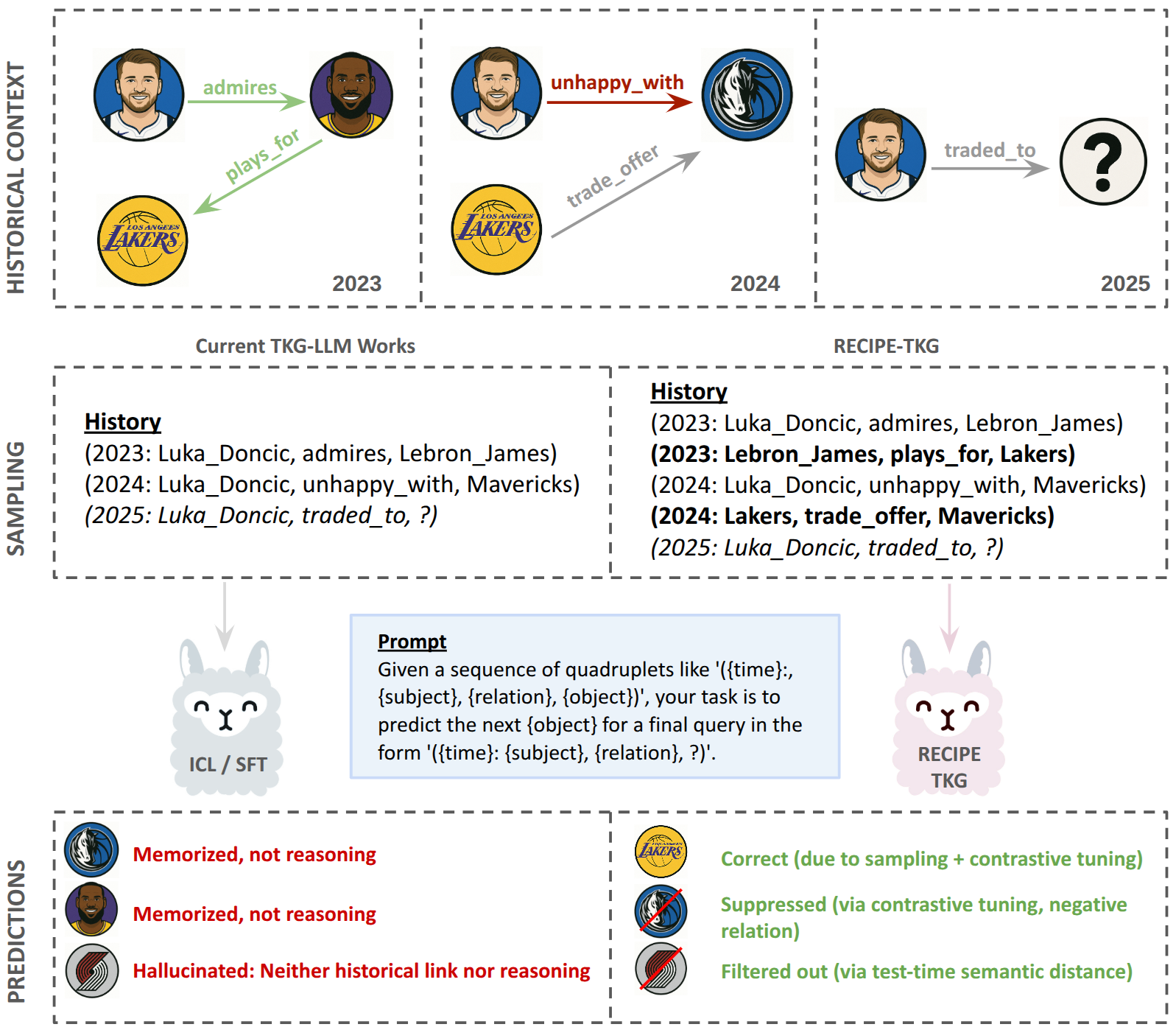}
    \caption{Example of LLM-based TKG reasoning. Prior methods rely on 1-hop historical context, leading to memorization or hallucination. RECIPE-TKG incorporates richer structural and relational context by sampling and filtering, enabling more plausible predictions.}
    \label{fig:example_tkg}
    \vspace{-5pt}
\end{figure}

Temporal Knowledge Graphs (TKGs) are widely used to represent dynamic, real-world knowledge across domains such as news~\cite{boschee2015icews, leetaru2013gdelt}, biomedicine~\cite{chaturvedi2024temporal}, and finance~\cite{dukkipati2025predictive}. They capture facts as time-stamped relational tuples (\texttt{subject}, \texttt{relation}, \texttt{object}, \texttt{timestamp}), modeling how interactions evolve over time~\cite{tresp2015learning}. A core task in this setting is TKG completion, which involves predicting missing or future links based on observed temporal interactions. This task requires reasoning over both relational and temporal structure, with downstream applications in forecasting and decision support~\cite{trivedi2017know, jin2019recurrent}.

The rise of Large Language Models (LLMs) has sparked interest in using pretrained generative models for TKG completion, driven by their generalization capability and emergent reasoning skills~\cite{liao2024gentkg, luo2024coh, lee2023tkgicl}. 
While LLM reasoning is often benchmarked on math or logic-based tasks~\cite{lewkowycz2022solving,wang2025tina}, TKG completion provides a complementary testbed that emphasizes two reasoning challenges: 1. Integrating temporal, structural, and relational information in the reasoning process, and 2. Relational generalization under sparse or indirect historical interactions.
% While LLM reasoning is often benchmarked on math or logic-based tasks~\cite{lewkowycz2022solving,wang2025tina}, TKG completion provides a complementary testbed that emphasizes temporal extrapolation, relational generalization, and reasoning under sparsity. It challenges models to reason over evolving relationships and produce semantically consistent predictions, making it a strong setting for studying structured reasoning in dynamic, knowledge-rich environments.
Recent prompting-based and fine-tuned LLM methods~\cite{lee2023tkgicl, liao2024gentkg, luo2024coh,xia2024chain} report promising results. However, closer inspection reveals that their predictions often reflect shallow pattern matching rather than deeper temporal or relational reasoning. As illustrated in Figure~\ref{fig:example_tkg}, these models frequently favor entities that are lexically similar or locally frequent in the input, even when more plausible completions exist based on the graph structure.

These limitations are particularly evident in sparse-context settings, where the query includes little prior interaction between the subject and potential target entities. In such cases, extrapolation from multi-hop or indirect paths is required. Without sufficient structural grounding, LLMs, whether zero-shot or fine-tuned, often produce predictions that are disconnected from the graph. Although standard metrics like Hits@k may improve, it is unclear whether such gains reflect true relational reasoning or memorization of shallow patterns. Moreover, prior work lacks systematic analysis across input regimes, especially for queries requiring generalization beyond observed history.

In this paper, we investigate how LLMs reason over temporal knowledge graphs. Our analysis shows that performance drops sharply when historical evidence is missing or structurally shallow, exposing a gap in current modeling approaches. To close this gap, we propose RECIPE-TKG, a LLM-based method consisting of three components:

%To address these issues, we propose RECIPE-TKG, a reasoning-centric framework designed to enhance both the predictive accuracy and semantic plausibility of LLM-based TKG completion. It consists of three components:
\begin{itemize}
\item \textbf{A rule-based multi-hop sampling strategy} that enriches the prompt with structurally and temporally diverse neighbors, providing better grounding for predictions.
\item \textbf{A contrastive fine-tuning objective} applied to lightweight adapter layers on a small subset of data, which shapes the embedding space around relational semantics.
\item \textbf{A semantic similarity-based filtering mechanism} that selects outputs at inference time using embedding proximity.
% inspired by the test-time compute (TTC) paradigm~\cite{snell2024ttc}.
\end{itemize}

Experiments on four benchmarks show that RECIPE-TKG outperforms previous LLM baselines with relative gains on Hits@1/3/10 ranging from 8\% to 30.6\% and returns more contextually plausible outputs, especially in challenging low-context settings. These results suggest that LLMs can be steered into more effective and reliable TKG forecasters when guided by the right context and training objectives.

\section{Challenges in TKG Completion with LLMs}
\label{section:challenges}

\begin{figure*}[ht]
    \vspace{-20pt}
    \centering
    \setlength{\abovecaptionskip}{5pt} 
    \setlength{\belowcaptionskip}{-8pt} 
    \includegraphics[width=\linewidth]{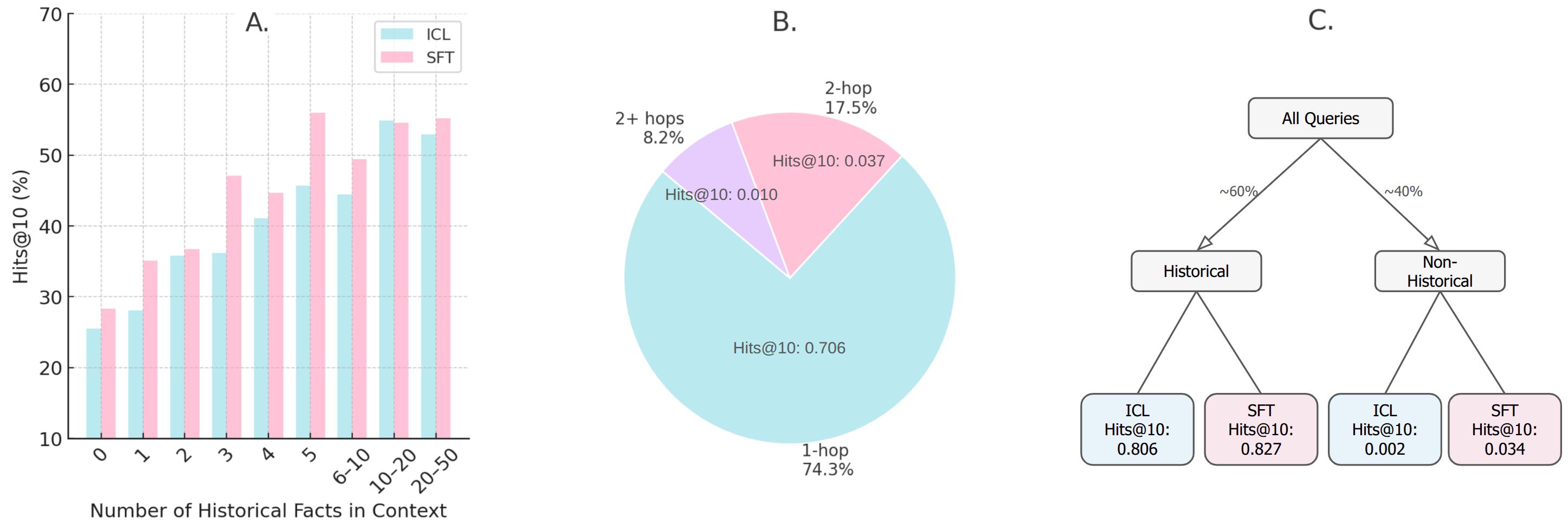}
    \caption{\textbf{Prediction failures under sparse or shallow history.} (a) Accuracy vs. history length shows longer contexts support better reasoning. (b) Most non-historical targets require multi-hop reasoning, but are unreachable with 1-hop sampling. (c) Accuracy drops sharply on non-historical predictions for both ICL and SFT.}
    \label{fig:grounding_failures}
\end{figure*}

Despite recent progress in adapting LLMs to TKG completion, these models often default to surface-level patterns in the input and fail to generate accurate predictions when structural or temporal cues are indirect, missing, or require multi-hop reasoning. To guide the design of our framework, we conduct a detailed empirical analysis of recent LLM-based approaches~\cite{lee2023tkgicl, liao2024gentkg} and examine some core challenges.
% two core challenges: grounding predictions in relevant historical context, and generalizing to queries where the correct answer is not directly observed. These observations motivate the components of RECIPE-TKG.

\subsection{Grounding Predictions in Historical Context}
\label{challenges:grounding}
Temporal Knowledge Graph completion requires models to reason over limited, evolving contexts. A key distinction in this setting is whether the correct object of a query has been observed in the sampled history. We define a prediction as \textit{historical} if the ground-truth entity appears in the retrieved context prior to the query time, and \textit{non-historical} otherwise. This distinction is crucial because existing LLM-based methods perform well on historical predictions but exhibit significant performance drops in non-historical cases, where memorization is insufficient and extrapolation is required.

% Figure~\ref{fig:grounding_failures}(a) shows that model accuracy 
% strongly correlates with the length of the input context. 
Figure~\ref{fig:grounding_failures}(a) shows that model performance improves with longer history: Hits@10 is below 0.3 with only one retrieved fact but exceeds 0.5 with 20–50 facts. This consistent trend for both in-context learning and fine-tuned models highlights the importance of providing sufficient historical evidence. 
% Figure~\ref{fig:grounding_failures}(b) shows that a significant portion of gold targets are not reachable via 1-hop from the query subject, making shallow sampling insufficient. In particular, over 25\% of queries require reasoning over 2-hop or more distant connections, and another 4\% are unreachable due to missing links. Models restricted to shallow views thus lack the structural context needed to ground the completions.
Figure~\ref{fig:grounding_failures}(b) further reveals that over 25\% of gold targets require multi-hop reasoning, while 4\% are unreachable due to missing links, making shallow sampling inadequate for many queries.

These effects are further amplified on non-historical predictions. As shown in Figure~\ref{fig:grounding_failures}(c), LLMs exhibit strong performance on historical predictions (e.g., 80–83\% Hits@10), as opposed to below 5\% when the target is non-historical. This gap reflects a reliance on lexical overlap or memorized associations, calling for a retrieval mechanism that recovers semantically and temporally relevant multi-hop context.

These findings motivate the first component of RECIPE-TKG: a rule-based, graph-aware multi-hop sampling strategy that retrieves structurally diverse and temporally aligned facts to support stronger contextual grounding, particularly for non-historical predictions.

\subsection{Limitations of Supervised Fine-Tuning}

Supervised fine-tuning (SFT) is widely used to adapt LLMs to TKG tasks, and prior work such as GenTKG~\cite{liao2024gentkg} reports notable improvements over prompting-based strategies~\citep{lee2023tkgicl}. However, our re-evaluation under controlled conditions shows that much of this improvement originates not from fine-tuning itself, but from differences in sampling strategies and evaluation setups.

\paragraph{Evaluation Frameworks Explain Much of the Gap.}
LLMs produce open-ended text that requires careful postprocessing to extract valid entity predictions. While \citet{lee2023tkgicl} uses a basic evaluation setup, GenTKG applies a more refined pipeline with canonicalization and output filtering, making direct comparisons misleading.

To disentangle these effects, we re-evaluate both prompting-based strategies and fine-tuned models with different sampling and evaluation pipelines under a unified framework. We compare naive sampling used in ~\citet{lee2023tkgicl} and TLR sampling ~\citep{liao2024gentkg}, and two evaluation settings (basic eval and GenTKG eval ~\citep{liao2024gentkg}). As shown in Table~\ref{tab:gentkg_vs_ours}, replacing the evaluation code alone increases Hits@1 from 25.8\% to 34.4\%. TLR sampling strategy adopted in GenTKG provides a modest improvement (35.1\%) compared to ICL sampling, while fine-tuning adds only a small additional gain (36.4\%). This suggests that a large portion of the reported gain stems from implementation choices, not from the model's improved reasoning capabilities.

\begin{table}[h]
\centering
\setlength{\abovecaptionskip}{5pt} 
\setlength{\belowcaptionskip}{-3pt}
\renewcommand{\arraystretch}{1.1}
\caption{Re-evaluation of ICL and SFT methods using consistent decoding and evaluation. The reported gains of GenTKG stem primarily from evaluation setup and sampling, with limited impact from fine-tuning.}
\resizebox{\linewidth}{!}{%
\begin{tabular}{lccc}
\toprule
\rowcolor{SuperSoftPink}
\textbf{Method} & \textbf{Hits@1} & \textbf{Hits@3} & \textbf{Hits@10} \\
\midrule
\multicolumn{4}{l}{\textit{Reported in GenTKG}} \\
\midrule
ICL (naive sampling + basic eval) & 0.258 & 0.430 & 0.510 \\
+ Fine-Tuning (TLR sampling + eval) & 0.369 & 0.480 & 0.535 \\
\midrule
\multicolumn{4}{l}{\textit{Re-evaluated under consistent setup}} \\
\midrule
ICL (naive sampling) + GenTKG eval & 0.344 & 0.464 & 0.523 \\
ICL (TLR sampling) + GenTKG eval & 0.351 & 0.473 & 0.527 \\
SFT (TLR sampling) + GenTKG eval & 0.364 & 0.476 & 0.532 \\
\bottomrule
\end{tabular}%
}
\label{tab:gentkg_vs_ours}
\vspace{-2mm}
\end{table}

\paragraph{Fine-Tuning Alone Does Not Improve Generalization.}
As established in Section~\ref{challenges:grounding}, both ICL and fine-tuned models struggle with non-historical predictions, where the correct answer does not appear in the retrieved history. These failures persist across a range of input sizes and are especially severe when the gold entity requires multi-hop reasoning, which is not supported by current sampling methods. Fine-tuning improves memorization of patterns seen during training but does not provide the relational inductive bias needed to reason about unseen or indirectly connected entities.

\paragraph{Motivating Contrastive Fine-Tuning.}
To address this limitation, we propose a contrastive fine-tuning objective that goes beyond correctness-based supervision. Rather than reinforcing output repetition, it explicitly trains the model to differentiate between semantically plausible and implausible candidates based on relational compatibility. In contrast to SFT, which rewards surface-level alignment with training data, contrastive learning reshapes the embedding space to support relational discrimination and generalization.

This motivates the second component of RECIPE-TKG: fine-tuning lightweight LoRA~\cite{hu2022lora} adapters using a contrastive objective to improve relational generalization and reduce hallucinations in sparse history settings.

% \subsection{Limitations of Exact-Match Metrics}

% Standard evaluation metrics for TKG completion, such as Hits@k, rely on exact string matching to determine correctness. While this approach is appropriate for strictly defined entity sets, it fails to reflect the semantic plausibility of predictions generated by LLMs, especially in open-ended generative settings.

% We observe that LLMs often produce outputs that are relationally appropriate but not identical to the ground truth. For example, when the correct answer is \texttt{Military\_Personnel\_(Nigeria)}, a model prediction of \texttt{Military\_(Nigeria)} may be semantically valid and even preferable under certain contexts, yet it is counted as incorrect under exact-match metrics.

% This mismatch is especially pronounced in sparse or non-historical settings, where the model is required to generalize from partial relational structure rather than copy memorized entities. Penalizing these semantically grounded outputs obscures the true reasoning ability of the model and biases evaluation toward lexical repetition.

% To address this, we introduce \textbf{Distance@k}, a metric that measures the semantic proximity between predicted and gold entities in embedding space. This allows us to capture predictions that are relationally coherent, even if not exact lexical matches, providing a more faithful measure of model plausibility and generalization.

\section{Preliminaries}

\paragraph{Problem Formulation.}
A Temporal Knowledge Graph is a collection of time-stamped facts represented as quadruples $(s, p, o, t)$, where $s$ and $o$ are subject and object entities, $p$ is a relation, and $t$ denotes the timestamp of the event. Formally, a TKG is denoted as $\mathcal{G} = (\mathcal{V}, \mathcal{R}, \mathcal{E}, \mathcal{T})$, where $\mathcal{V}$ is the set of entities, $\mathcal{R}$ the relations, $\mathcal{E}$ the event facts, and $\mathcal{T}$ the time indices. Each time step $t$ defines a historical snapshot $\mathcal{G}_t \subseteq \mathcal{E}$.

The forecasting task involves predicting a missing entity in a future quadruple. Given a query of the form $(s, p, ?, t)$ or $(?, p, o, t)$ and a set of historical snapshots $\{\mathcal{G}_1, \ldots, \mathcal{G}_{t-1}\}$, the model must return the most plausible entity that completes the query at time $t$.

\paragraph{Low-Rank Adaptation (LoRA)}
To reduce the number of trainable parameters, we adopt LoRA~\cite{hu2022lora}, which re-parameterizes the weight update as 
{\setlength{\abovedisplayskip}{5pt}%
 \setlength{\belowdisplayskip}{5pt}%
 \begin{equation}
 \hat{h}(x) = W_0 x + A B x,
 \end{equation}}

where $W_0$ is a frozen pretrained weight and $A$, $B$ are trainable low-rank matrices.

\section{Method}

\begin{figure*}[ht]
    \vspace{-30pt}
    \centering
    \setlength{\abovecaptionskip}{5pt} 
    \setlength{\belowcaptionskip}{-5pt} 
    \includegraphics[trim=25 50 25 30, clip, width=\linewidth]{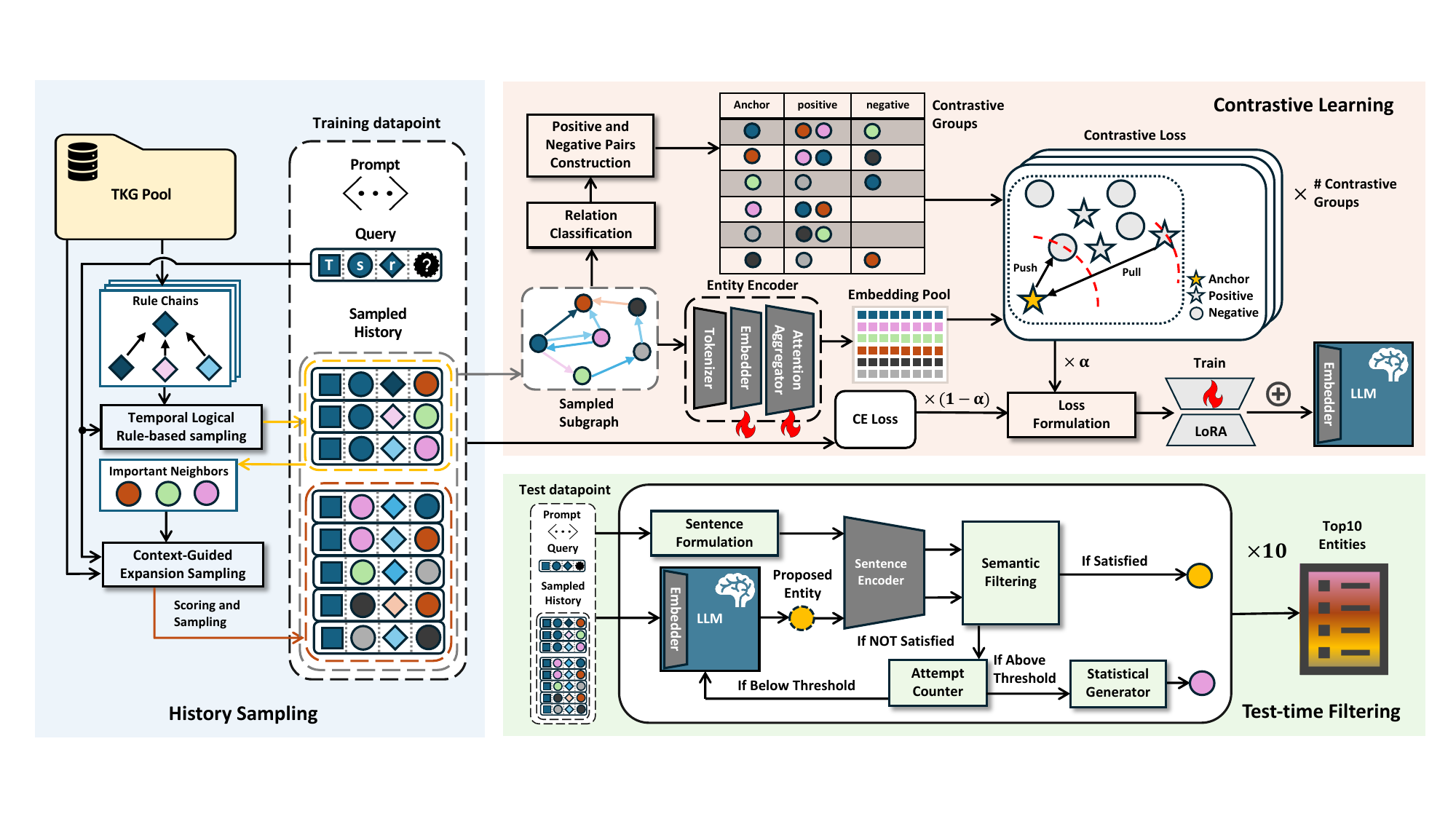}
    \caption{
    % Overview of RECIPE-TKG. RECIPE-TKG consists of three stages: (1) \textbf{History Sampling}, which retrieves query-relevant historical facts using a two-phase approach combining temporal logical rule-based sampling and context guided expansion sampling to ensure structural relevance and semantic diversity; (2) \textbf{Contrastive Learning}, where entity embeddings are optimized via a hybrid loss combining contrastive and cross-entropy objectives. Positive and negative entity pairs are constructed from the sampled subgraph, and entity embeddings are encoded by a learnable entity encoder composed of a tokenizer, an embedder, and an attention aggregator. The LLM is fine-tuned using LoRA; and (3) \textbf{Test-time Filtering}, where proposed entities generated by the LLM are iteratively verified through a semantic filtering module. If the output is unsatisfactory, alternative candidates are generated using a statistical generator. This process repeats until top-ranked predictions are produced.
    \textbf{Overview of RECIPE-TKG.} RECIPE-TKG follows a three-stage framework: (1) \textbf{History Sampling}, which retrieves query-relevant facts via a two-phase strategy combining rule-based retrieval and context-guided expansion; (2) \textbf{Contrastive Learning}, which jointly optimizes entity embeddings using contrastive and cross-entropy losses. Positive/negative pairs are sampled from the subgraph, and embeddings are generated via a learnable encoder; (3) \textbf{Test-time Filtering}, where predicted entities are iteratively verified by a semantic filter. Unsatisfactory outputs are refined using a statistical generator until confident predictions are obtained.
}
    \label{fig:framework}
\end{figure*}

% In this section, we present RECIPE-TKG, a three-stage framework designed to improve the predictive accuracy and semantic plausibility of LLM-based temporal knowledge forecasting.

In this section, we present RECIPE-TKG, a three-stage LLM-based framework for temporal knowledge forecasting. The complete framework is illustrated in Figure~\ref{fig:framework}. 

% It addresses the limitations identified in Section~\ref{section:challenges} by: (1) retrieving historically and structurally relevant context through a multi-hop sampling strategy that combines temporal rule-based retrieval with graph-aware expansion; (2) applying a contrastive fine-tuning objective that encourages the model to distinguish plausible from implausible entity completions based on relational structure; and (3) refining generated outputs at inference time using a semantic similarity-based filtering mechanism. 
% Our complete framework is illustrated in Figure~\ref{fig:framework}. 

% Together, these components enable more reliable predictions, especially in sparse or non-historical settings where standard prompting or fine-tuning fails.

% \subsection{Rule-Based Multi-hop History Sampling Strategy / Structure-aware RAG variant over TKGs}
\subsection{RBMH: Rule-Based Multi-Hop History Sampling}
\label{sec:sampling}
The first stage of RECIPE-TKG focuses on retrieving a compact yet informative history from the temporal knowledge graph $\mathcal{G}$. For a given query $(s_q, p_q, ?, T)$, we aim to retrieve historical facts $\{(s, p, o, t) \in \mathcal{G} \mid t < T\}$ that are temporally valid and structurally relevant. Our sampling process combines rule-based retrieval with context-guided expansion to provide richer support for reasoning, particularly in sparse or non-historical settings.

\paragraph{Stage 1: Temporal Logical Rule-based Sampling.}

We begin by retrieving subject-aligned 1-hop facts using a rule-based procedure adapted from TLR~\citep{liao2024gentkg}, which learns relational rules of the form $p_q \Leftarrow \{p_{b_1}, \dots, p_{b_k}\}$ through 1-step temporal random walks, capturing event regularities. We retrieve historical quadruples $(s, p, o, t)$ such that $s = s_q$ and $p$ appears in the rule body for the query relation $p_q$. See Appendix~\ref{appendix:history sampling} for the details. 
% This yields a high-precision local context that emphasizes temporal coherence and predicate-level patterns

However, this 1-hop retrieval cannot reach facts involving semantically relevant but structurally distant entities. Due to the fixed number of learned rules, this stage often retrieves fewer than $N$ quadruples, the maximum the LLM can handle. This motivates a second stage to expand context with more diverse and informative facts.

\paragraph{Stage 2: Context-guided Multi-hop Expansion} 
We then samples additional historical facts from $\mathcal{G}$. The candidate pool includes any quadruples not retrieved in Stage 1 whose subjects differ from $s_q$.

This stage is designed to support multi-hop reasoning by identifying facts that may not directly connect to the query subject but are structurally and semantically relevant. Each candidate $(s, p, o, t)$ is assigned a composite weight:
\begin{equation}
w = w_n \cdot w_f \cdot (w_t + w_c + w_{cp}),
\label{eq:weight}
\end{equation}
where \(w_n\) downweights unreachable or distant nodes, \(w_f\) penalizes high-frequency triples, \(w_t\) prioritizes temporal recency, \(w_c\) favors co-occurrence with the query subject or relation, and \(w_{cp}\) reinforces connectivity with the initial TLR context. 
% These components guide sampling toward facts that facilitate deeper relational inference while preserving temporal and structural plausibility.

To sample from candidate pool, We first select the top $10 \times M$ candidates by score to form a reduced pool. From this pool, we sample $M$ quadruples with probabilities proportional to their weights. This soft filtering strategy preserves diversity while prioritizing high-quality candidates, avoiding over-reliance on only the highest-scoring facts. Our two-stage RBMH sampling method supports reasoning beyond immediate neighbors and avoids overfitting to shallow or overly common facts. The overall design motivation, formal definitions, hyperparameters and algorithms are provided in Appendix~\ref{appendix:weights}.

\subsection{Contrastive Fine-Tuning for Structured Reasoning}
\label{sec:contrastive}
To improve generalization beyond memorized entity associations, we introduce a contrastive fine-tuning objective that supplements the standard next-token prediction loss, helping to disambiguate plausible from implausible predictions, especially when historical context is sparse or indirect.

% This component encourages the model to learn entity-level distinctions based on their relational context, 

% Our approach involves three steps: (1) constructing contrastive groups using labeled relation types, (2) learning entity embeddings via token-level attention, and (3) jointly optimizing contrastive and generative objectives.

\paragraph{Relation-Guided Contrastive Pair Construction.}
Our design is guided by the international relations principle, \textit{The enemy of my enemy is my friend}, which reflects relational patterns common in geopolitical TKGs and motivates how we position entities in embedding space. Inspired by this structure, we first categorize relations into \textbf{positive}, \textbf{negative}, and \textbf{neutral} types using \texttt{GPT-4o}, minimizing the inclusion of neutral cases (see Appendix~\ref{appendix:relation}). Given a sampled subgraph (Figure~\ref{fig:framework}), we treat each unique entity as an anchor and examine its 1-hop neighbors. A neighbor is assigned as a \textit{positive} sample if it connects via a positive relation, or a \textit{negative} sample if it connects via a negative relation. If both types of edges exist, the neighbor is excluded to avoid contradiction. Neutral relations are ignored. This process forms contrastive groups that are used to calculate the contrastive loss. 

\paragraph{Entity Embedding Encoding.}
Since an entity typically spans multiple tokens, we adopt a multi-stage process to compute its representation. First, the entity string is tokenized. Each resulting token is then passed through the model’s embedding layer (embedder), which produces an embedding vector. These token embeddings $\{h_1, h_2, \ldots, h_k\}$ are subsequently aggregated into a single entity-level embedding $e$ using a trainable \textbf{attention aggregator}.

The final embedding is a weighted sum:
\vspace{-0.4em}
\begin{equation}
e = \sum_{j=1}^{k} \lambda_j h_j,
\end{equation}
where $\lambda_j$ are attention weights satisfying $\sum_j \lambda_j = 1$. Both the embedding layer and the aggregator are learnable modules, jointly optimized during fine-tuning.

\paragraph{Training Objective.}
The overall loss function is defined as:
% \vspace{-0.2em}
\begin{equation}
\mathcal{L} = \alpha \cdot 
\mathcal{L}_{\text{contrastive}} + (1 - \alpha) \cdot \mathcal{L}_{\text{ce}}(o, o_p),
\end{equation}
% \vspace{-0.2em}
where $\mathcal{L}_{\text{ce}}$ denotes the cross-entropy loss between the predicted token $o_p$ and the ground truth $o$, $\mathcal{L}_{\text{contrastive}}$ represents the contrastive loss, and $\alpha \in [0, 1]$ is a balancing hyperparameter.

The contrastive loss is formulated as:
\vspace{-0.4em}
\begin{multline}
\mathcal{L}_{\text{contrastive}} = \frac{1}{N_c} \sum_{i=1}^{N_c} \max\Big(0, \\
\|a_i - pos_i\|^2 - \|a_i - neg_i\|^2 + m\Big)
\end{multline}

where \( N_c \) is the number of contrastive groups, and \( a_i \) denotes the embedding of the anchor entity. For each group, \( pos_i \) is the hardest positive, defined as the farthest positive entity from the anchor in the embedding space, while \( neg_i \) is the closest negative. This formulation emphasizes challenging examples and enforces a margin \( m \) to improve the separation between positive and negative pairs.

This training objective encourages the model to pull the most distant positive samples closer to the anchor and push the nearest negatives farther away. This dynamic adjustment refines the semantic structure of the latent space, enabling better entity discrimination and improving downstream reasoning performance. More details can be found at Appendix~\ref{appendix:train}.
% This dynamic adjustment refines the semantic structure of the latent space, 

\subsection{Similarity-Based Test-Time Filtering}
\label{sec:filtering}

Recent work shows that language models can improve inference without parameter updates by using lightweight test-time strategies~\cite{snell2024ttc, ji2025testtime}. Building on this idea, we introduce a semantic similarity-based filtering method to reduce hallucinations by removing predictions misaligned with the input context.

% Recent studies have shown that language models can improve inference performance without modifying model parameters by leveraging lightweight test-time strategies~\cite{snell2024ttc, ji2025testtime}. Inspired by this idea, we introduce a semantic similarity-based filtering mechanism that post-processes the model’s predictions to mitigate hallucinations, i.e., predictions that are semantically misaligned with the input context or unjustifiably novel.

\begin{figure}
    
    \centering
    \setlength{\abovecaptionskip}{5pt} 
    \setlength{\belowcaptionskip}{-3pt} 
    \includegraphics[width=\linewidth]{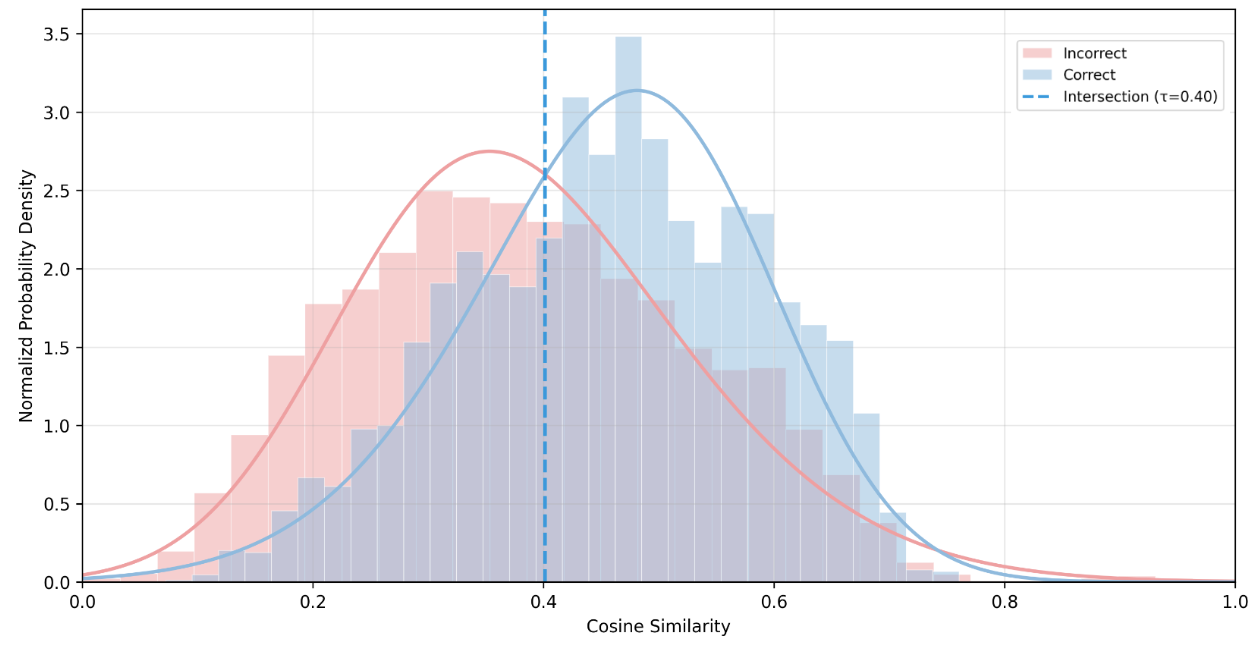}
    \caption{Distribution of semantic similarity values for correctly and incorrectly classified samples to the input context.}
    \label{fig:dist}
    \vspace{-2mm}
\end{figure}

Our filtering approach is motivated by two empirical observations: 
\begin{enumerate}
    \item Models often generate non-historical entities that have low semantic alignment with the input context, especially in sparse settings despite higher similarity scores correlating with correctness (Figure~\ref{fig:dist}).
    \item In many cases, the ground truth entity already appears in the historical context $\mathcal{H}$, yet the model produces a non-historical prediction that yields negligible gain in accuracy.
\end{enumerate}

These patterns suggest that enforcing semantic consistency and reconsidering salient entities from the input can correct many low-quality predictions. Rather than rejecting or reranking predictions with fixed rules, we apply an adaptive refinement strategy grounded in semantic similarity.

\vspace{0.5em}
\paragraph{Semantic Consistency Verification.}
We embed the generated prediction $p$ and the input context $c$ using a sentence transformer model to compute a similarity score:
\begin{align}
\phi(p,c) &= \text{cos-sim}(E(p), E(c)) \\
E(x) &= \text{SentenceTransformer}(x) \in \mathbb{R}^{d}
\end{align}
where $E(\cdot)$ denotes the output vector of a pretrained transformer model. We use this similarity as a proxy for contextual alignment. A prediction is accepted if its similarity score exceeds a learned threshold $\tau$, or if it already appears in the retrieved history $\mathcal{H}$. Otherwise, we regenerate until a satisfactory prediction is found, or fall back to history-aware scoring.

This process is formalized as:
\begin{equation}
p^\prime = 
\begin{cases}
p & \text{if } p \in \mathcal{H} \ \text{or} \ \phi(p,c) \geq \tau \\
\text{regenerate}(p) & \text{if } \phi(p,c) < \tau \\
\arg\max_{h\in\mathcal{H}} \psi(h) & \text{after } k \text{ attempts}
\end{cases}
\end{equation}

Figure~\ref{fig:framework} illustrates how filtering interacts with the generation process to improve robustness.

\vspace{0.5em}
\paragraph{Historical Relevance Fallback.}
If repeated generations yield unsatisfactory predictions, we fall back to the historical candidates $\mathcal{H}$. Each candidate $h \in \mathcal{H}$ is scored by:
\begin{equation}
\psi(h) = \beta \cdot f(h) + (1 - \beta) \cdot r(h)
\end{equation}
where $f(h)$ is the frequency of $h$ in the input history and $r(h)$ captures recency. This mechanism biases the selection toward historically grounded entities when semantic alignment fails.

\vspace{0.5em}
\paragraph{Threshold Selection.}
The threshold $\tau$ is optimized to best separate correct and incorrect predictions based on empirical distributions of $\phi(p, c)$. We describe the optimization objective and quantitative justification in Appendix~\ref{appendix:filtering}, along with implementation details such as embedding model choice and fallback hyperparameters.

\section{Experiments}

\subsection{Experimental Setup}

\paragraph{Proposed method.}
We refer to our full method as RECIPE-TKG, which combines rule-based multi-hop history sampling (\textit{RBMH Sampling}), contrastive fine-tuning denoted as \textit{CFT}, and \textit{Test-time Filtering}. 
% All variants by substituting one or several of the components are implemented on top of the same base language models for fair comparison.

\paragraph{Language Models.} 
% Our primary experiments are conducted on \texttt{LLaMA-2-7B}~\cite{touvron2023llama2}, a widely used open-source LLM. To ensure modern relevance, we also evaluate \texttt{LLaMA-3-8B}~\cite{meta2024llama3}, both of which demonstrate improved instruction-following capabilities and alignment. These models are used in both in-context learning and fine-tuning setups under consistent prompt structures, with further training and filtering details provided in Appendix~\ref{apppendix:training_details}.

Our primary experiments are conducted on \texttt{LLaMA-2-7B}~\cite{touvron2023llama2}, a widely used open-source model in LLM-based TKG completion research~\cite{liao2024gentkg,luo2024coh}. To ensure modern relevance, we also evaluate \texttt{LLaMA-3-8B}~\cite{meta2024llama3}. Prompts and implementation details are provided in Appendix~\ref{apppendix:prompt} and~\ref{apppendix:training_details}

\paragraph{Datasets.} 
% We evaluate our method on four widely-used benchmark datasets: \textbf{ICEWS14}, \textbf{ICEWS18}, \textbf{GDELT}, and \textbf{YAGO}. ICEWS datasets~\cite{boschee2015icews} contain timestamped political events such as (``Angela Merkel'', Visit, ``India'', 2015-03-25). GDELT~\cite{leetaru2013gdelt} includes global socio-political interactions over time. YAGO~\cite{mahdisoltani2013yago3} comprises a curated subset of facts with temporal annotations. Each dataset contains queries of the form $(s, p, ?, t)$ or $(?, p, o, t)$ for evaluation. Detailed dataset statistics can be found in Appendix~\ref{sec:dataset_stats}.

We evaluate RECIPE-TKG on four commonly adopted benchmark datasets: 
ICEWS14 and ICEWS18, both derived from the ICEWS project~\cite{boschee2015icews}, 
GDELT~\cite{leetaru2013gdelt}, and 
YAGO~\cite{mahdisoltani2013yago3}. 
Detailed dataset statistics are provided in Appendix~\ref{appendix:dataset_stats}.

\paragraph{Evaluation Metrics.} 
We choose temporal-aware filtered Hits@1/3/10 as our evaluation metrics, following prior work~\cite{gastinger2023comparing}. 
% To better capture the reasoning quality of LLMs, we introduce a semantic-aware metric, \textbf{Distance@k}, which quantifies the semantic similarity between predicted and ground-truth entities using sentence transformer embeddings. This metric helps assess the plausibility of predictions beyond exact string match and is described further in Appendix~\ref{sec:similarity_metric}.

\paragraph{Baselines.} 
We compare RECIPE-TKG against three categories of methods. \textbf{Embedding-based methods} include RE-GCN~\cite{li2021temporal}, xERTE~\cite{han2020xerte}, TANGO~\cite{han2021learning}, and TimeTraveler~\cite{sun2021timetraveler}. \textbf{Rule-based method} includes TLogic~\cite{liu2022tlogic}. Finally, \textbf{LLM-based methods} include ICL ~\cite{lee2023tkgicl} and GenTKG~\cite{liao2024gentkg}. Additional information about baseline methods is included in Appendix~\ref{appendix:baseline_details}.

\subsection{Main Results}
\label{sec:main_results}

\begin{table*}[t]
\vspace{-10pt}
\centering
\setlength{\abovecaptionskip}{5pt} 
\setlength{\belowcaptionskip}{-3pt} 
\renewcommand{\arraystretch}{1.2}
\setlength{\tabcolsep}{6pt}
\caption{\textbf{Temporal link prediction results on temporal-aware filtered Hits@1/3/10.} LLM-based models are implemented based on \texttt{LLaMA2-7B}. Best results for each metric are highlighted in \textbf{bold}, and the best results among LLM-based models are \underline{underlined}. The last row shows the relative improvement (\(\Delta\)) of RECIPE-TKG over the best-performing LLM-based baseline.}

\resizebox{\textwidth}{!}{
\begin{tabular}{ll|ccc|ccc|ccc|ccc}
\toprule
\rowcolor{SuperSoftPink}
\multicolumn{2}{c|}{\textbf{Datasets}} & 
\multicolumn{3}{c|}{\textbf{ICEWS14}} & 
\multicolumn{3}{c|}{\textbf{ICEWS18}} & 
\multicolumn{3}{c|}{\textbf{GDELT}} & 
\multicolumn{3}{c}{\textbf{YAGO}} \\
\rowcolor{SuperSoftPink}
\multicolumn{2}{c|}{\textbf{Models}} & 
Hits@1 & Hits@3 & Hits@10 & 
Hits@1 & Hits@3 & Hits@10 & 
Hits@1 & Hits@3 & Hits@10 & 
Hits@1 & Hits@3 & Hits@10 \\

\midrule

\multirow{4}{*}{\centering Embedding-based} 
& RE-GCN~\cite{li2021temporal}         & 0.313 & 0.473 & 0.626 & 0.223 & 0.367 & \textbf{0.525} & 0.084  & 0.171 & 0.299 & 0.468 & 0.607 & 0.729 \\
& xERTE~\cite{han2020xerte}            & 0.330 & 0.454 & 0.570 & 0.209 & 0.335 & 0.462 & 0.085  & 0.159 & 0.265 & 0.561 & 0.726 & 0.789 \\
& TANGO~\cite{han2021learning}         & 0.272 & 0.408 & 0.550 & 0.191 & 0.318 & 0.462 & 0.094  & 0.189 & 0.322 & 0.566 & 0.651 & 0.718 \\
& Timetraveler~\cite{sun2021timetraveler} & 0.319 & 0.454 & 0.575 & 0.212 & 0.325 & 0.439 & 0.112 & 0.186 & 0.285 & 0.604 & 0.770 & 0.831 \\
\midrule

\multirow{1}{*}{\centering Rule-based} 
& TLogic~\cite{liu2022tlogic}          & 0.332 & 0.476 & 0.602 & 0.204 & 0.336 & 0.480 & \textbf{0.113} & \textbf{0.212} & \textbf{0.351} & 0.638 & 0.650 & 0.660 \\
\midrule

\multirow{3}{*}{\centering LLM-based} 
& ICL~\cite{lee2023tkgicl}            & 0.344 & 0.464 & 0.523 & 0.164 & 0.302 & 0.382 & 0.090 & 0.172 & 0.242 & 0.738 & 0.807 & 0.823 \\
& GenTKG~\cite{liao2024gentkg}        & 0.364 & 0.476 & 0.532 & 0.200 & 0.329 & 0.395 & \underline{0.099} & \underline{0.193} & 0.280 & 0.746 & 0.804 & 0.821 \\
& \textbf{RECIPE-TKG}                 & \underline{\textbf{0.393}} & \underline{\textbf{0.526}} & \underline{\textbf{0.651}} & \underline{\textbf{0.224}} & \underline{\textbf{0.369}} & \underline{0.516} & 0.095 & 0.192 & \underline{0.327} & \underline{\textbf{0.811}} & \underline{\textbf{0.880}} & \underline{\textbf{0.930}} \\
\cmidrule(lr){3-14}
\rowcolor{white}
& $\Delta$ 
& \cellcolor{LightGray}8.0\% 
& \cellcolor{LightGray}10.5\% 
& \cellcolor{LightGray}22.4\% 
& \cellcolor{LightGray}12.0\% 
& \cellcolor{LightGray}12.2\% 
& \cellcolor{LightGray}30.6\% 
& \cellcolor{LightGray}{-4.0\%} 
& \cellcolor{LightGray}{-0.5\%} 
& \cellcolor{LightGray}16.8\% 
& \cellcolor{LightGray}8.7\% 
& \cellcolor{LightGray}9.0\% 
& \cellcolor{LightGray}13.0\% \\
% \rowcolor{LightGray}
% & $\Delta$ & +8.0\% & +10.5\% & +22.4\% & +12.0\% & +12.2\% & +30.6\% & -4.0\% & -0.5\% & +16.8\% & +8.7\% & +9.0\% & +13.0\% \\
\bottomrule
\end{tabular}
}
\label{tab:combined_results}
\vspace{-2mm}
\end{table*}

% Results from Table~\ref{tab:combined_results} demonstrate that RECIPE-TKG has strong performance across the four benchmarks, surpassing both embedding-based and LLM-based baselines on nearly all evaluation metrics. On ICEWS14 and YAGO, RECIPE-TKG achieves state-of-the-art performance with up to 11.9\% relative improvement on the best baselines respectively. On ICEWS18, RECIPE-TKG outperforms the best LLM-based baseline by 30.6\% relative improvement on Hits@10. It also has comparable results with RE-GCN ,the best embedding-based method on ICEWS18. While RECIPE-TKG does not outperform the rule-based method of TLogic on GDELT, it still achieves the best Hits@10 (32.7\%) among all LLM-based models and maintains comparable results on Hits@1 and Hits@3. The strong performance of RECIPE-TKG further positions LLM-based models as strong candidates for foundation models in temporal knowledge graph completion.

Results in Table~\ref{tab:combined_results} show that RECIPE-TKG consistently performs well across four benchmark datasets, surpassing both embedding-based and LLM-based baselines on nearly all evaluation metrics. On ICEWS14 and YAGO, RECIPE-TKG establishes new state-of-the-art results, achieving up to 11.9\% relative improvement over the strongest competing methods. For ICEWS18, it exceeds the best LLM-based baseline by a substantial margin, with a 30.6\% relative gain in Hits@10, and achieves comparable performance to RE-GCN, the top embedding-based approach on this dataset. Although RECIPE-TKG does not outperform the rule-based method TLogic on GDELT, it attains the highest Hits@10 score (32.7\%) among all LLM-based models and remains competitive on Hits@1 and Hits@3. These results highlight the effectiveness of RECIPE-TKG and further positions LLM-based methods as strong candidates for foundation models in temporal knowledge graph completion.

\section{Analysis}
\label{sec:analysis}

% We now present a comprehensive evaluation of RECIPE-TKG across four temporal knowledge graph forecasting benchmarks. Our results demonstrate that the proposed framework not only improves standard accuracy metrics such as Hits@k, but also better addresses the reasoning challenges outlined in Section~\ref{section:challenges}, including prediction under sparsity, generalization to non-historical entities, and hallucination mitigation.

\subsection{Ablation Study}
\label{sec:ablation}

% \begin{table*}[t]
% \centering
% \renewcommand{\arraystretch}{1.15}
% \setlength{\tabcolsep}{5.5pt}
% \caption{\textbf{Ablation study on ICEWS14 with \texttt{LLaMA2-7B}.} We evaluate in-context learning, supervised fine-tuning (SFT), and contrastive fine-tuning (Contrastive FT) across three history sampling strategies.}
% \resizebox{\textwidth}{!}{
% \begin{tabular}{ll|ccc|ccc|ccc}
% \toprule
% \rowcolor{SuperSoftPink}
% \multicolumn{2}{c|}{\textbf{Model}} & \multicolumn{3}{c|}{ICL} & \multicolumn{3}{c|}{SFT} & \multicolumn{3}{c}{Contrastive FT} \\
% \rowcolor{SuperSoftPink}
% \multicolumn{2}{c|}{} & Hits@1 & Hits@3 & Hits@10 & Hits@1 & Hits@3 & Hits@10 & Hits@1 & Hits@3 & Hits@10 \\
% \midrule
% \multirow{3}{*}{\centering Sampling}
% & \citet{lee2023tkgicl} & 0.344 & 0.464 & 0.523 & 0.360 & 0.469 & 0.530 & 0.363 & 0.479 & 0.529 \\
% & TLR - \citet{liao2024gentkg} & 0.351 & 0.473 & 0.527 & 0.364 & 0.476 & 0.532 & 0.367 & 0.476 & 0.532 \\
% & RBMH (Ours)  & 0.364 & 0.500 & 0.572 & 0.389 & 0.519 & 0.582 & 0.392 & 0.521 & 0.580 \\
% \bottomrule
% \end{tabular}
% }
% \label{tab:ablation_icews14}
% \vspace{-2mm}
% \end{table*}

\begin{table}[t]
\centering
\setlength{\abovecaptionskip}{5pt} 
\setlength{\belowcaptionskip}{-3pt} 
\renewcommand{\arraystretch}{1.15}
\setlength{\tabcolsep}{1.5pt}
\caption{\textbf{Ablation study on ICEWS14 with \texttt{LLaMA2-7B}.} Comparison of training paradigms across different history sampling strategies. The bold results show the original combinations of components in prior works and our method.}
\tiny
\begin{tabular}{l|ccc|ccc|ccc}
\toprule
\rowcolor{SuperSoftPink}
& \multicolumn{3}{c|}{ICL} & \multicolumn{3}{c|}{SFT} & \multicolumn{3}{c}{CFT} \\
\rowcolor{SuperSoftPink}
& H@1 & H@3 & H@10 & H@1 & H@3 & H@10 & H@1 & H@3 & H@10 \\
\midrule
~\citet{lee2023tkgicl}     & \textbf{0.344} & \textbf{0.464} & \textbf{0.523} & 0.360 & 0.469 & 0.530 & 0.363 & 0.479 & 0.529 \\
TLR~\cite{liao2024gentkg}  & 0.351 & 0.473 & 0.527 & \textbf{0.364} & \textbf{0.476} & \textbf{0.532} & 0.367 & 0.476 & 0.532 \\
RBMH                         & 0.364 & 0.500 & 0.572 & 0.389 & 0.519 & 0.582 & \textbf{0.392} & \textbf{0.521} & \textbf{0.580} \\
\bottomrule
\end{tabular}
\label{tab:ablation_icews14_small}
\vspace{-2mm}
\end{table}

We did ablation studies of the framework to individually evaluate the effects of key components in our framework, \textit{RBMH Sampling} and \textit{CFT}, against the components in prior works. Component combinations are denoted using hyphenated labels. We first conduct a structured ablation study on ICEWS14 using \texttt{LLaMA2-7B}. We compare three sampling methods (~\citet{lee2023tkgicl}, TLR~\cite{liao2024gentkg}, and our proposed \textit{RBMH Sampling}) and three training paradigms: in-context learning (ICT), supervised fine-tuning (SFT) and our proposed contrastive fine-tuning (CFT). As shown in Table~\ref{tab:ablation_icews14_small}, the bold results highlight original combinations from prior works and RECIPE-TKG w/o filtering, while the rest show disentangled individual contributions of each component. The results demonstrate that \textit{RBMH Sampling} consistently improves performance across all training paradigms, demonstrating the benefit of retrieving structurally diverse and semantically relevant context. While \textit{CFT} performs comparably to SFT under the same sampling strategy, it exhibits clear advantages when historical context is sparse or entirely absent. As discussed in Appendix~\ref{appendix:contrastive_analysis}, contrastive models generate predictions that are semantically closer to the ground truth, even when exact matches are not possible. This suggests that contrastive objectives promote structure-aware generalization beyond surface-level accuracy, especially in sparse or non-historical settings where lexical cues are insufficient.

% The findings reveal that our history sampling method consistently enhances performance across all supervision strategies, underscoring the importance of retrieving structurally diverse and semantically relevant context. Second, contrastive fine-tuning does not provide more substantial gains than standard fine-tuning when evaluated under the same sampling strategy. For example, under our sampling, CL reaches 0.580 Hits@10 compared to FT’s 0.582 and ICL’s 0.572. However, contrastive tuning offers distinct advantages in settings where historical context is sparse or entirely absent. To better understand this effect, we analyze how contrastive tuning impacts prediction quality beyond exact-match metrics like Hits@k. As detailed in Appendix~\ref{appendix:contrastive_analysis}, we measure the embedding-space distance between predicted and ground-truth entities across varying history lengths. This analysis shows that contrastive models consistently produce semantically closer predictions under limited history, where other models often fail. Even when the gold entity is not explicitly observed in the input, contrastively trained models return alternatives that are more relationally compatible—revealing a form of structure-aware generalization not captured by surface-level accuracy metrics. These results highlight that while retrieval quality sets an upper bound on LLM performance, contrastive tuning enables better use of retrieved context—particularly in sparse and non-historical settings where lexical cues are insufficient. 

\begin{table}[t]
\vspace{-8pt}
\centering
\setlength{\abovecaptionskip}{5pt} 
\setlength{\belowcaptionskip}{-3pt} 
\renewcommand{\arraystretch}{1.15}
\setlength{\tabcolsep}{5pt}
\caption{Effect of removing RECIPE-TKG components.}
\resizebox{\columnwidth}{!}{
\begin{tabular}{l|ccc}
\toprule
\rowcolor{SuperSoftPink}
\textbf{SETTINGS} & Hits@1 & Hits@3 & Hits@10 \\
\midrule
RECIPE-TKG w/o CFT & 0.364 & 0.501 & 0.643 \\
RECIPE-TKG w/o RBMH Sampling & 0.364 & 0.483 & 0.581 \\
RECIPE-TKG w/o Filtering & 0.392 & 0.521 & 0.580 \\
RECIPE-TKG & 0.393 & 0.526 & 0.651 \\
\bottomrule
\end{tabular}
}
\label{tab:ablation2}
\vspace{-1mm}
\end{table}

Table~\ref{tab:ablation2} provides additional insights into the effects of each of the three components, especially \textit{test-time filtering}. When comparing the CFT-RBMH setting with and without \textit{Test-time Filtering}, we observe a substantial boost in Hits@10 from 0.580 to 0.651, underscoring the effectiveness of our test-time refinement mechanism. Notably, combining \textit{test-time filtering} with \textit{RBMH Sampling} and \textit{Test-time Filtering} (\textbf{RECIPE-TKG}) yields the best performance across all metrics. 

\subsection{Performance Gains Across Input Regimes}

\begin{figure}[t]
\vspace{-2pt}
\centering
\setlength{\abovecaptionskip}{5pt} 
\setlength{\belowcaptionskip}{-3pt} 
  \includegraphics[width=\linewidth]{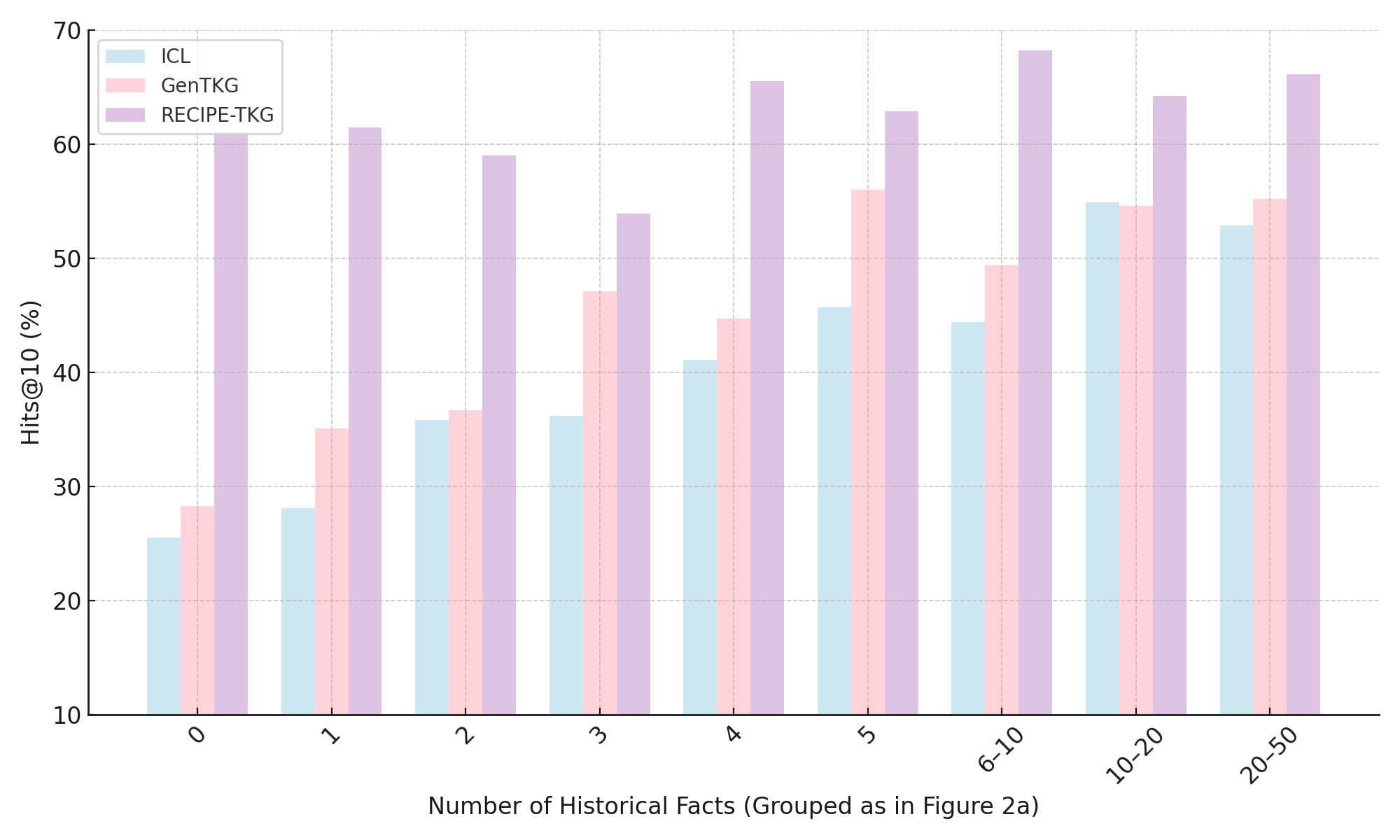}
\caption{Hits@10 grouped by number of historical facts. RECIPE-TKG consistently outperforms ICL and GenTKG across all history lengths, with particularly strong improvements when the input history is sparse.}
  \label{fig:history_bins}
  \vspace{-2mm}
\end{figure}

To evaluate how historical input affects model performance, we group queries by the number of retrieved facts and compare Hits@10 across methods. These bins align with Figure~\ref{fig:grounding_failures}(a), allowing direct comparison with prior failure patterns. As shown in Figure~\ref{fig:history_bins}, RECIPE-TKG outperforms both ICL and GenTKG across all groups, with especially large gains in the low-history regime.

Two key insights emerge. First, prior failures on short-history queries were not due to intrinsic difficulty, but rather to shallow retrieval. Since all methods are evaluated on the same query set, the strong gains from RECIPE-TKG (reaching over 60\% Hits@10 for history length 0 to 2) indicate that even sparse queries can be completed accurately when provided with deeper, multi-hop context. This validates the effectiveness of \textit{RBMH Sampling} in recovering structurally and temporally relevant support.

Second, RECIPE-TKG continues to outperform baselines even with longer histories (10–50 facts), where other methods begin to plateau. This sustained advantage reflects the contributions of \textit{CFT} and \textit{Test-time Filtering}, which improve generalization and reduce hallucinations.

Overall, these results show that RECIPE-TKG not only addresses the limitations of shallow context but also improves reasoning and prediction quality across a wide range of query types.

\subsection{Case Study: Performance of Llama3-8b}

As shown in Table~\ref{tab:llama2_vs_llama3}, \texttt{LLaMA3-8B} performs comparably to \texttt{LLaMA2-7B}, supporting our choice of the latter for most experiments. Moreover, this choice of base model enables a fair comparison with prior work using fine-tuned models.
Under both backbones, RECIPE-TKG consistently outperforms ICL, demonstrating its robustness and generalizability across different LLMs. 
% Moreover, this marginal improvement on \texttt{LLaMA3-8B}  helps mitigate the concerns regarding information leakage~\citep{sannidhi2024retrieval,ding2024zrllm,xia2024chain} when testing recent LLMs on established benchmarks like ICEWS14.

\begin{table}[t]
\centering
\renewcommand{\arraystretch}{1.2}
\setlength{\tabcolsep}{6pt}
\caption{Comparison between \texttt{LLaMA2-7B} and \texttt{LLaMA3-8B} on ICEWS14.}
\resizebox{\columnwidth}{!}{
\begin{tabular}{l|ccc|ccc}
\toprule
\rowcolor{SuperSoftPink}
\textbf{Model} & \multicolumn{3}{c|}{\textbf{LLaMA2-7B}} & \multicolumn{3}{c}{\textbf{LLaMA3-8B}} \\
\rowcolor{SuperSoftPink}
& hit@1 & hit@3 & hit@10 & hit@1 & hit@3 & hit@10 \\
\midrule
ICL       & 0.344 & 0.464 & 0.523  & 0.351 & 0.484 & 0.578 \\
RECIPE-TKG          & 0.393 & 0.526 & 0.651  & 0.367 & 0.529 & 0.658 \\
\bottomrule
\end{tabular}
}
\label{tab:llama2_vs_llama3}
\end{table}

% \subsection{Discussion}

% Taken together, these experiments validate our core hypothesis: improving LLM-based TKG completion requires not just stronger architectures or larger models, but a principled integration of retrieval, supervision, and refinement mechanisms. RECIPE-TKG systematically addresses the limitations outlined in Section~\ref{section:challenges}, particularly the difficulty of non-historical reasoning, the limited utility of surface-level fine-tuning, and the inadequacy of exact-match metrics. By aligning history sampling with temporal structure, reshaping the embedding space around relational semantics, and applying lightweight semantic feedback during decoding, our method achieves both higher accuracy and more plausible predictions across domains.
\section{Conclusion}
\label{sec:conclusion}

We introduced RECIPE-TKG, a framework for improving LLM-based temporal knowledge graph forecasting through rule-based multi-hop history sampling, contrastive fine-tuning, and test-time filtering. Our results show consistent improvements in accuracy and robustness, particularly in sparse or non-historical settings where previous methods fail to generalize. By aligning retrieved context with relational structure and refining predictions at inference time, RECIPE-TKG advances structured reasoning in generative models without requiring large-scale retraining. This highlights the potential of modular strategies to adapt LLMs for temporally grounded reasoning on knowledge graphs.

\section*{Limitations}

Although RECIPE-TKG adopts a structured three-stage framework, it is still built on clean, fully observed temporal knowledge graphs, which may not reflect real-world scenarios. The rule mining step requires offline learning before sampling, and must be repeated if the TKG changes. Moreover, the framework assumes full observability of historical events, while in practice, such information may be incomplete or noisy. Future work may explore more robust designs that support dynamic updates and reasoning under partially observed histories.

\section*{License and Ethics}

% All datasets used in this study are publicly available and licensed for academic research. LLaMA-2 and LLaMA-3 models are used under Meta's research license. No personal or sensitive data is used in training or evaluation. We adhere to the ethical standards of the datasets and models employed.

All datasets used in this study are publicly available and licensed for academic research. Specifically, ICEWS14, ICEWS18, GDELT, and YAGO have been widely adopted in prior work on temporal knowledge graphs. No personally identifiable information (PII) or sensitive content is present in any of the datasets.

We use LLaMA-2 and LLaMA-3 models under Meta’s official research license, and all model adaptations are conducted in compliance with their intended use for academic and non-commercial research. The training and evaluation procedures are entirely conducted on benchmark data, and no human subjects are involved.

We adhere to the ethical guidelines set forth by the ACL Code of Ethics, including transparency, reproducibility, and the responsible use of language models. Our work poses minimal risk of harm and does not involve content generation, human annotation, or interaction with real users.

\clearpage
\bibliographystyle{acl_natbib}
%\bibliography{anthology}

\clearpage
\appendix
\section{Rule-Based Multi-Hop History Sampling Details}
\subsection{TLR Algorithm}
\label{appendix:history sampling}

Algorithm~\ref{alg:tlr} shows the TLR retrieval procedure used in our framework, reproduced from~\cite{liao2024gentkg}.

\begin{algorithm}[h]
\caption{TLR Retrieval}
\label{alg:tlr}
\textbf{Input:} Temporal knowledge graph $\mathcal{G}$, query $(s_q, r_q, ?, T)$, learned rules $\mathcal{TR}$\\
\textbf{Output:} A set of retrieved facts $\mathcal{G}_{s_q}(s_q, r_q, T)$
\begin{algorithmic}[1]
\State $\mathcal{G}_{s_q}(s_q, r_q, T) \gets \emptyset$
\For{$fact \gets (s_q, r_q, o, t<T)) \in \mathcal{G}$}
    \State $\mathcal{G}_{s_q}(s_q, r_q, T) \gets \mathcal{G}_{s_q}(s_q, r_q, T) \cup fact$
\EndFor
\For{\textbf{top k rules} $w.r.t.\ r_q \gets r_b \in \mathcal{TR}$}
    \State Get a list $r_b \gets \{r_{b_1}, r_{b_2}, \cdots, r_{b_k}\}$
\EndFor
\For{$fact \gets (s_q, r \in r_b, o, t<T) \in \mathcal{G}$}
    \State $\mathcal{G}_{s_q}(s_q, r_q, T) \gets \mathcal{G}_{s_q}(s_q, r_q, T) \cup fact$
\EndFor
\State \textbf{return} $\mathcal{G}_{s_q}(s_q, r_q, T)$
\end{algorithmic}
\end{algorithm}

\subsection{Context-guided Multi-hop Expansion Details}
\label{appendix:weights}
\subsubsection{Weight Formulation Discussion}
\label{appendix:weight-discussion}

We adopt a multiplicative combination of the weight components rather than a simple sum to for two reasons. First, the neighbor weight \( w_n \) acts as a hard constraint: it equals zero if the subject or object of a candidate quadruple is not reachable from the query, effectively filtering out irrelevant facts. Second, the frequency weight \( w_f \) is designed to down-weight commonly repeated triples while preserving their relative order. This logarithmic scaling ensures that rare but structurally relevant facts are not overshadowed. Together, the multiplicative form enables a soft prioritization across dimensions while preserving hard structural constraints.
\subsubsection{Weight Component}
The five weight components of equation~\ref{eq:weight} are defined as follows:

\textbf{Neighbor weight} $w_n$ ensures that structurally closer quadruples receive higher scores:
\[
w_n = \exp\left(-\gamma_1 \cdot (\text{hop}_s + \text{hop}_o - 1)\right),
\]
where $\text{hop}_s$ and $\text{hop}_o$ denote the shortest hop distances from the subject and object to the query subject. The weight decays exponentially with increasing distance, and vanishes to zero when either $\text{hop}_s$ or $\text{hop}_o$ is infinite, corresponding to cases where the entity is not reachable from the query subject in the graph. Importantly, all structural statistics (e.g., hop distance, co-occurrence counts, and context connectivity) are computed over the subgraph excluding quadruples with timestamps after the query time \( T \).

\textbf{Frequency weight} $w_f$ reduces the dominance of frequent triples (history quadruples excluding timestamp):
\[
w_f = \frac{1}{\gamma_2 \cdot \log(n_{spo}) + 1},
\]
where $n_{spo}$ is the count of the subject-predicate-object triple. This logarithmic form discourages over-sampling of repetitive patterns while maintaining frequency order.

More precisely, for any two triples with frequency counts \( n_1 < n_2 \), the corresponding weights satisfy:
\[
w(n_1) > w(n_2), \quad \text{and} \quad \frac{w(n_1)}{w(n_2)} = \frac{\log(n_2) + 1}{\log(n_1) + 1},
\]
assuming all other components of the weight function are equal. This shows that the multiplicative formulation preserves the relative ranking induced by frequency, while still suppressing the absolute dominance of highly frequent triples.

\textbf{Time weight} \( w_t \) favors temporally recent events:
\[
w_t = \exp\left( -\gamma_3 \cdot \frac{T - t}{\delta} \right),
\]
where \( T \) is the timestamp of the query, \( t \) is the timestamp of the event quadruple (with \( T > t \)), \( \delta \) is the time granularity (e.g., \( \delta = 24 \) in ICEWS14), and \( \gamma_3 \) controls the decay rate.

\textbf{Connection weight} $w_c$ promotes inclusion of frequently co-occurring entity pairs:
\[
w_c = \frac{\log(1 + \gamma_4 \cdot n_{so})}{1 + \log(1 + \gamma_4 \cdot n_{so})},
\]
where $n_{so}$ is the co-occurrence count of the subject-object pair prior to $T$, and $\gamma_4$ is a smoothing parameter. This bounded function emphasizes structural relevance while limiting hub bias.

\textbf{Contextual priority weight} $w_{cp}$ encourages sampling quadruples that remain connected to the initial TLR sampled subgraph:
\[
w_{cp} =
\begin{cases}
1, & \text{if } s \in \mathcal{E}_{\text{TLR}} \text{ or } o \in \mathcal{E}_{\text{TLR}}, \\
0, & \text{otherwise},
\end{cases}
\]
where $\mathcal{E}_{\text{TLR}}$ is the set of all 1-hop neighbors identified in the TLR stage. This guides the expansion toward semantically coherent subgraphs.

% \medskip

% Overall, each weight $w_x \in [0,1]$ is interpretable and controlled by hyperparameters $\gamma_1$, $\gamma_2$, $\gamma_3$, and $\gamma_4$. In our setting, $\gamma_1=0.6$, $\gamma_2=0.6$, $\gamma_3=0.01$ and $\gamma_4=0.1$.

\subsubsection{Hyperparameter Sensitivity Experiment}

\begin{figure}[t]
    \centering
    \includegraphics[width=0.9\linewidth, trim=14 20 16 10, clip]{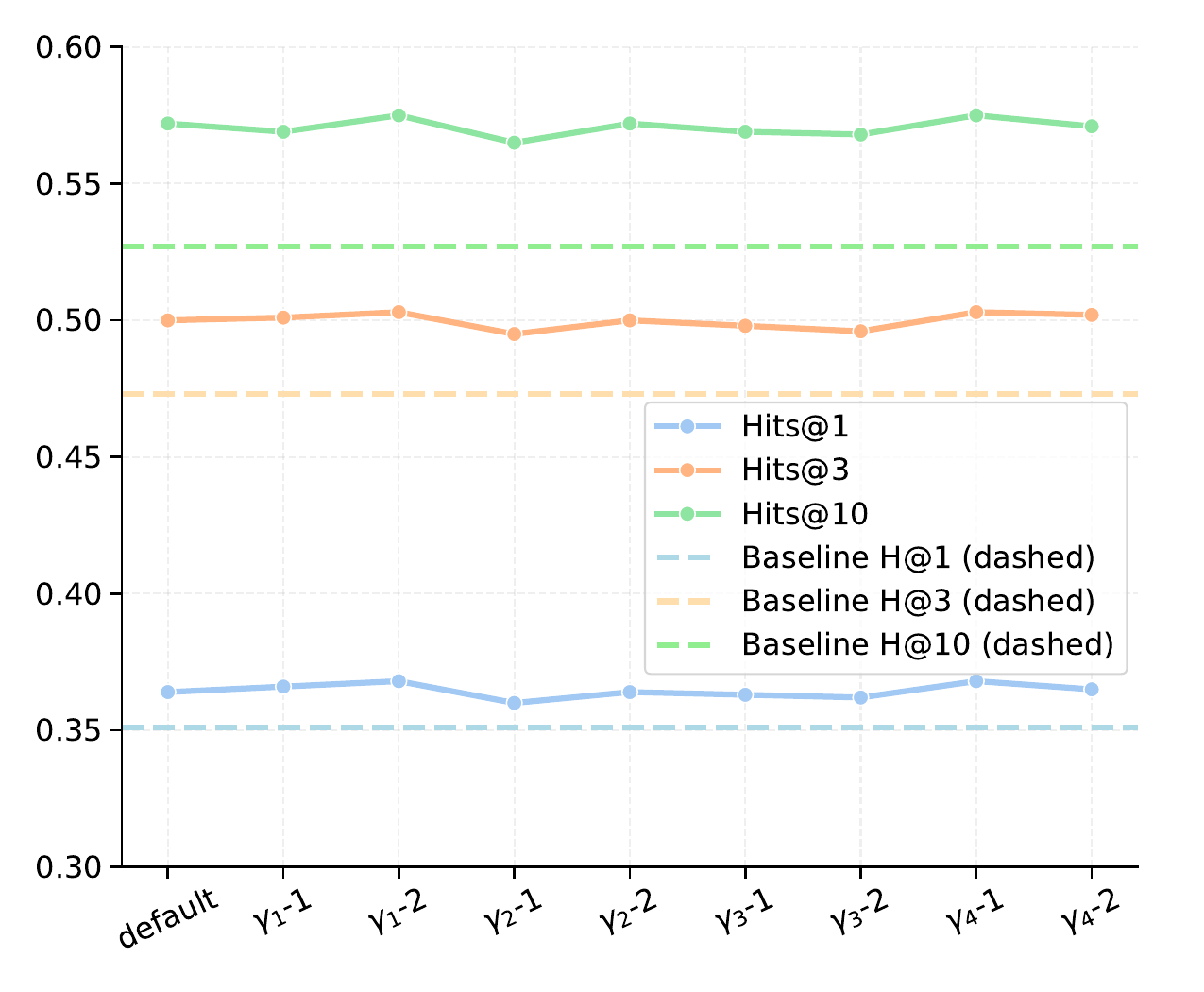}
    \caption{Performance of ICL-RBMH under different sampling hyperparameter configurations.}
    \label{fig:sampling_sensitivity}
\end{figure}

Figure~\ref{fig:sampling_sensitivity} presents the performance in ICL-RBMH setting under varying sampling hyperparameters. We perturb each of the four $\gamma_i$ parameters individually (two settings per parameter), while keeping others fixed, and compare them against the default configuration. Across all variants, model performance remains stable, indicating that \textit{RBMH Sampling} is robust to hyperparameter choices. Moreover, ICL-RBMH consistently outperforms the baseline ICL-TLR across all settings. 

The sampling hyperparameter configurations and their corresponding performance metrics are summarized in Table~\ref{tab:appendix_hyperparam}, including mean and standard deviation to reflect stability.

\begin{table}[ht]
\centering
\footnotesize
\setlength{\tabcolsep}{3.5pt}
\caption{Performance of ICL-RBMH under different sampling hyperparameter configurations on ICEWS14.}
\begin{tabular}{c|cccc|ccc}
\toprule
ID & $\gamma_1$ & $\gamma_2$ & $\gamma_3$ & $\gamma_4$ & Hits@1 & Hits@3 & Hits@10 \\
\midrule
default         & 0.6 & 0.6 & 0.01  & 0.1  & 0.364 & 0.500 & 0.572 \\
$\gamma_1$-1    & 0.4 & 0.6 & 0.01  & 0.1  & 0.366 & 0.501 & 0.569 \\
$\gamma_1$-2    & 0.8 & 0.6 & 0.01  & 0.1  & 0.368 & 0.504 & 0.575 \\
$\gamma_2$-1    & 0.6 & 0.4 & 0.01  & 0.1  & 0.364 & 0.500 & 0.572 \\
$\gamma_2$-2    & 0.6 & 0.8 & 0.01  & 0.1  & 0.364 & 0.500 & 0.572 \\
$\gamma_3$-1    & 0.6 & 0.6 & 0.05  & 0.1  & 0.363 & 0.498 & 0.569 \\
$\gamma_3$-2    & 0.6 & 0.6 & 0.002 & 0.1  & 0.368 & 0.506 & 0.573 \\
$\gamma_4$-1    & 0.6 & 0.6 & 0.01  & 0.2  & 0.368 & 0.503 & 0.575 \\
$\gamma_4$-2    & 0.6 & 0.6 & 0.01  & 0.05 & 0.365 & 0.502 & 0.571 \\
\midrule
\multicolumn{5}{c|}{Mean} & 0.366 & 0.501 & 0.571 \\
\multicolumn{5}{c|}{Std}  & 0.0020 & 0.0024 & 0.0021 \\
\midrule
\multicolumn{5}{c|}{Baseline (ICL-TLR)} & 0.351 & 0.473 & 0.527 \\
\bottomrule
\end{tabular}
\label{tab:appendix_hyperparam}
\end{table}

\subsection{RBMH Algorithm}
\begin{algorithm}[h]
\caption{Rule-based Multi-hop history sampling}
\label{alg:rbmh}
\textbf{Input:} Temporal knowledge graph $\mathcal{G}$, query $(s_q, r_q, ?, T)$, learned rules $\mathcal{TR}$, maximum history length $N$, scoring function $\mathcal{F}$, a set of TLR retrieved facts $\mathcal{G}_{s_q}(s_q, r_q, T)$\\
\textbf{Output:} A set of retrieved facts $\mathcal{G}(s_q, r_q, T)$
\begin{algorithmic}[1]
\State $M \gets$ $N - $ $len(\mathcal{G}_{s_q}(s_q, r_q, T))$ 
\If{$M = 0$}
    \State $\mathcal{G}(s_q, r_q, T) \gets$ $\mathcal{G}_{s_q}(s_q, r_q, T)$
    \State \textbf{return} $\mathcal{G}(s_q, r_q, T)$
\EndIf

\State $\mathcal{C} \gets \{(s, r, o, t, \mathcal{F}(s, r, o, t)) \mid (s, r, o, t) \in \mathcal{G},\ t < T \}$
\State $\mathcal{C}_{\text{top}} \gets \text{Top}_{10M}(\mathcal{C})$
\State $\mathcal{C}_{\text{sample}} \gets \text{WeightedSample}(\mathcal{C}_{\text{top}},\ M)$
\State $\mathcal{G}_{\text{mh}}(s_q, r_q, T) \gets \{(s, r, o, t)\ |\ (s, r, o, t, w) \in \mathcal{C}_{\text{sample}} \}$
\State $\mathcal{G}(s_q, r_q, T) \gets \mathcal{G}_{s_q}(s_q, r_q, T) \cup \mathcal{G}_{\text{mh}}(s_q, r_q, T)$
\State \Return $\mathcal{G}(s_q, r_q, T)$
\end{algorithmic}
\end{algorithm}

\section{Training Details}
\label{appendix:train}
\subsection{Relation Classification}
\label{appendix:relation}
The prompt used for relation classification is provided in Figure~\ref{fig:relation_prompt}.

\begin{figure*}[t]
\centering
\begin{tcolorbox}[
  colback=white, colframe=black, coltitle=white,
  title=Prompt for Relation Classification, fonttitle=\bfseries,
  colbacktitle=black, arc=3mm, boxrule=0.8pt,
  enhanced, unbreakable
]
You are analyzing relation labels from a political event knowledge graph, where each relation reflects an action or request within a geopolitical context.

Classify the sentiment of the given relation as one of the following:

\begin{itemize}
  \item \textbf{positive} (e.g., promoting peace, aid, cooperation)
  \item \textbf{negative} (e.g., violence, repression, aggression)
  \item \textbf{neutral} (e.g., procedural or ambiguous actions)
\end{itemize}

Avoid selecting \texttt{"neutral"} unless the relation is genuinely ambiguous or purely procedural in nature.
\end{tcolorbox}
\caption{Prompt used for relation classification.}
\label{fig:relation_prompt}
\end{figure*}

In cases where a neighbor is connected to the anchor via both a positive and a negative relation, it is excluded in training to avoid ambiguity.

\begin{figure}[t]
  \centering
  \includegraphics[trim=6 5 10 10, clip, width=\linewidth]{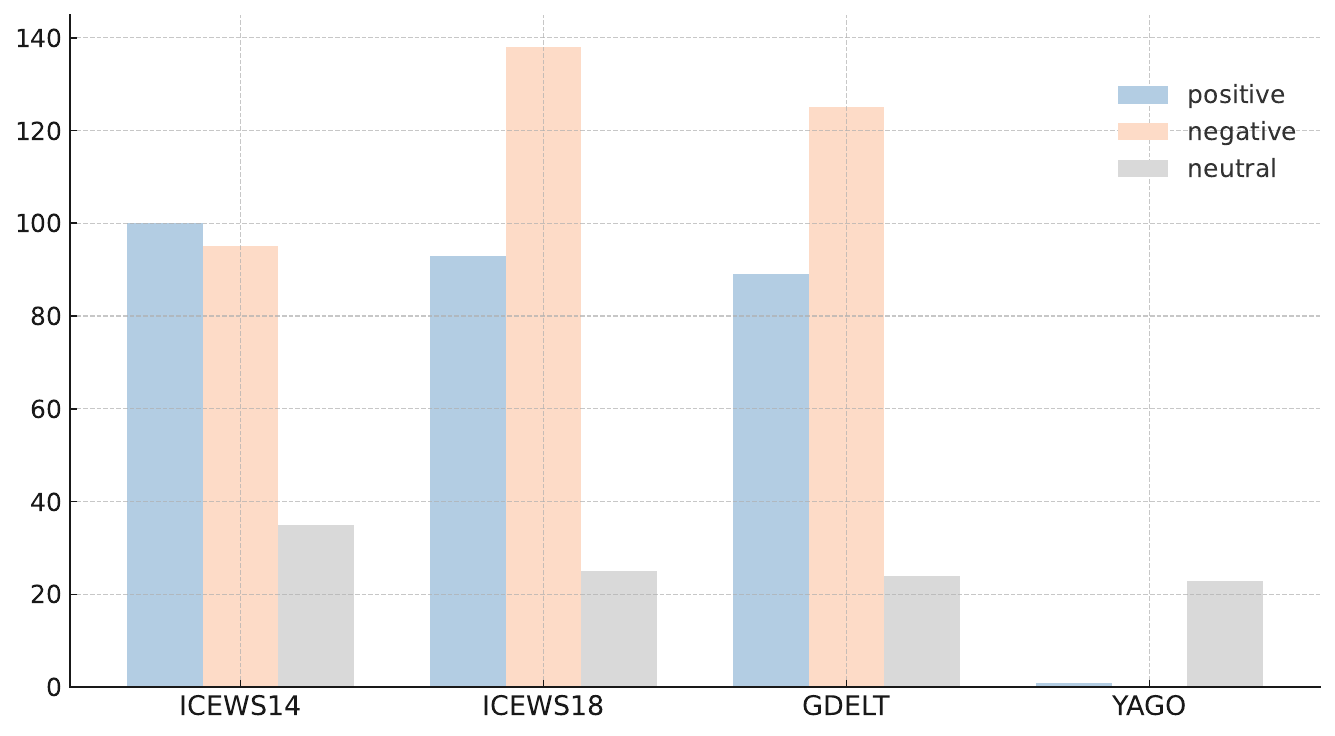}
  \caption{Distribution of relation types in four datasets after automatic classification.}
  \label{fig:relation_type_distribution_adjusted_legend}
\end{figure}

Figure~\ref{fig:relation_type_distribution_adjusted_legend} shows the distribution of relation types across four datasets. Positive and negative relations appear in roughly balanced proportions, while neutral relations are consistently less common. Notably, YAGO exhibits a distinct relation distribution where the majority of relations are classified as \textit{neutral}. Upon inspection, we find that this reflects the actual semantic nature of the relations in the dataset, which are mostly descriptive or taxonomic rather than sentiment-oriented. Consequently, the contrastive learning component has limited impact on YAGO, as it relies on meaningful distinctions between positive and negative relations. The observed performance gain on YAGO is therefore primarily attributed to improvements in history sampling and \textit{Test-time filtering}.

% \subsection{Entity encoding}
% \label{appendix:entity_encoding}

% Under our contrastive learning objective, the entity embeddings produced by this aggregation module are refined in latent space. Specifically, entities involved in semantically positive interactions are pulled closer to each other, while those involved in negative interactions are pushed apart. This joint training encourages the aggregator to produce entity representations that are not only context-aware, but also semantically discriminative.

\subsection{Prompt}
\label{apppendix:prompt}
To guide the language model in performing temporal knowledge completion, we adopt a structured, instruction-style prompt format shown in Figure~\ref{fig:prompt}. The prompt defines the task explicitly: given a chronological list of historical events represented as quadruples, the model must predict the missing object entity for a future temporal query.

% Each historical fact is formatted as "\texttt{{time}:[{subject}, {relation}, {object_label}.{object}]}", where \texttt{\{object\_label\}} is a unique identifier associated with the entity (e.g., \texttt{3380.Joseph\_Robinette\_Biden}). This labeling scheme facilitates consistent reference resolution and improves post-processing via regex-based extraction. The final input ends with the query, and the model is asked to generate the correct object in fully qualified form \texttt{\{object\_label\}.\{object\}}.

Each historical fact is formatted as \texttt{\detokenize{{time}:[{subject}, {relation}, {object_label}.{object}]}} where \texttt{\{object\_label\}} is a unique identifier associated with the entity (e.g., \texttt{3380.Joseph\_Robinette\_Biden}). This labeling scheme facilitates consistent reference resolution and improves post-processing via regex-based extraction. The final input ends with the query, and the model is asked to generate the correct object in fully qualified form \texttt{\{object\_label\}.\{object\}}.

This prompt format is applied consistently across both in-context learning and fine-tuning setups.

\begin{figure*}[t]
\centering
\begin{tcolorbox}[
  colback=white, colframe=black, coltitle=white,
  title=Prompt Example, fonttitle=\bfseries,
  colbacktitle=black, arc=3mm, boxrule=0.8pt,
  enhanced, unbreakable
]
You must be able to correctly predict the next \texttt{\{object\}} from a given text consisting of multiple quadruplets in the form of "\texttt{\{time\}:[\{subject\}, \{relation\}, \{object\_label\}.\{object\}]}" and the query in the form of "\texttt{\{time\}:[\{subject\}, \{relation\},}" in the end. You must generate \texttt{\{object\_label\}.\{object\}}.

\medskip

\begin{flushleft}
\texttt{%
2014-01-15: [Mehmet\_Simsek, Make\_statement, 5195.Other\_Authorities\_(Turkey)] \\
2014-01-20: [Nuri\_al-Maliki, Consult, 3380.Joseph\_Robinette\_Biden] \\
2014-01-25: [Joseph\_Robinette\_Biden, Make\_an\_appeal, 3990.Massoud\_Barzani] \\
2014-02-01: [Joseph\_Robinette\_Biden, Make\_an\_appeal\_or\_request,}
\end{flushleft}

\end{tcolorbox}
\caption{Instruction-style prompt format for TKG forecasting.}
\label{fig:prompt}
\end{figure*}

\subsection{LoRA Formulation}

We follow the standard LoRA setup~\cite{hu2022lora}. Given a frozen pretrained weight matrix $W_0 \in \mathbb{R}^{d \times k}$, LoRA introduces two trainable low-rank matrices $A \in \mathbb{R}^{d \times r}$ and $B \in \mathbb{R}^{r \times k}$ with $r \ll \min(d, k)$, such that the original forward transformation $h(x) = W_0 x$ is modified as:
\begin{equation}
\hat{h}(x) = W_0 x + ABx.
\end{equation}
This design allows efficient fine-tuning by only training $A$ and $B$, while keeping the pretrained weights $W_0$ frozen. In our experiments, we adopt the default LoRA implementation from the PEFT library~\cite{peft2022}.

\subsection{Implementation Details}
\label{apppendix:training_details}
% \textit{
% This section is a placeholder. We need to update.}
We fine-tune \texttt{LLaMA-2-7B} and \texttt{LLaMA-3-8B} models using LoRA adapters. All trainings are conducted on 2 H100 GPUs in \texttt{bfloat16} precision. We set maximum history length to 50 in history sampling according to the context length of \texttt{LLaMA-2-7B}. For fine-tuning, we train 1024-shots data for 50 epochs with the batch size of 512, the learning rate of 3e-4, the context length of 4096, the target length of 128, the LoRA rank of 8, the LoRA dropout rate of 0.05. For RECIPE-TKG, we train 6024-shots data (1024 aligned with GenTKG and 5000 randomly sampled by seed 42) for 10 epochs, and other settings keep unchanged. Contrastive tuning uses a margin of 1.0 and loss weight $\alpha = 0.2$ to balance cross-entropy and contrastive objectives.

Entities are tokenized using the native tokenizer of the LLM and embedded via the model’s embedding layer. A lightweight attention aggregator produces final entity embeddings, jointly trained with the model. 

\subsection{Hyperparameter Sensitivity Experiment}

\begin{figure}[t]
  \centering
  \includegraphics[trim=6 5 10 10, clip, width=\linewidth]{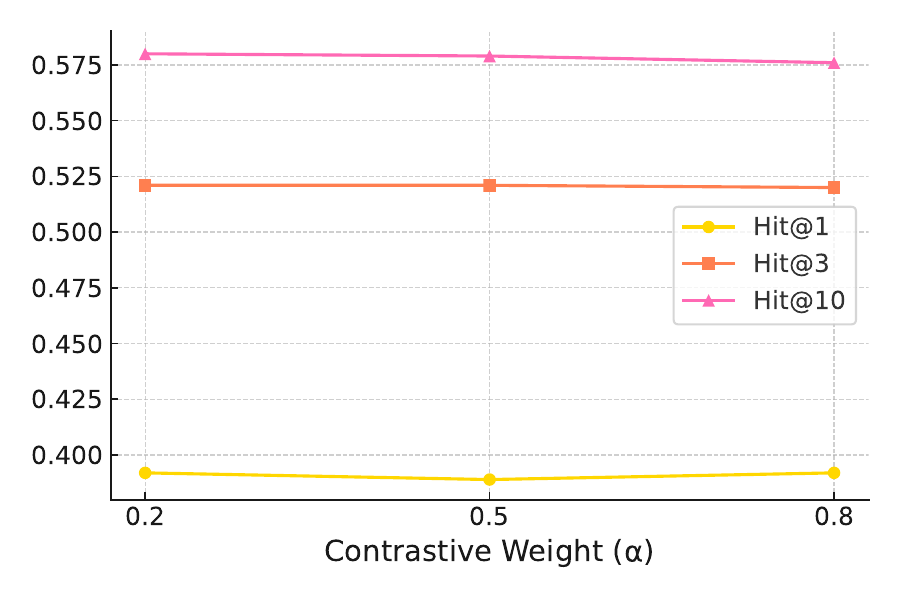}
  \caption{Effect of contrastive weight ($\alpha$)}
  \label{fig:contrastive_weight}
\end{figure}

As shown in Figure~\ref{fig:contrastive_weight}, varying $\alpha$ from 0.2 to 0.8 leads to marginal fluctuations across all evaluation metrics. These results suggest that the model is robust to the choice of $\alpha$, and that \textit{CFT} contributes consistently across a wide range of weighting schemes. Table~\ref{tab:contrastive_weight} presents the sensitivity of model performance to the contrastive weight $\alpha$. The consistently small standard deviations across metrics suggest that the model is robust to variations in $\alpha$.

\begin{table}[t]
\centering
\caption{Performance under different contrastive weight settings on ICEWS14.}
\label{tab:contrastive_weight}
\begin{tabular}{lccc}
\toprule
Weight $\alpha$ & Hits@1 & Hits@3 & Hits@10 \\
\midrule
0.2 & 0.392 & 0.521 & 0.580 \\
0.5 & 0.389 & 0.521 & 0.579 \\
0.8 & 0.392 & 0.520 & 0.576 \\
\midrule
Mean & 0.391 & 0.521 & 0.578 \\
Std  & 0.0014 & 0.0006 & 0.0020 \\
\bottomrule
\end{tabular}
\end{table}

\section{Test-Time Filtering}
\label{appendix:filtering}

\paragraph{Embedding Model.}
To compute semantic similarity between predictions and context, we use the \texttt{all-mpnet-base-v2} model~\cite{song2020mpnet,all-mpnet-base-v2} from HuggingFace, a pretrained sentence transformer with 768-dimensional output. We treat both the generated prediction string and the full in-context prompt as input sequences and extract mean-pooled embeddings for similarity calculation.

\paragraph{Similarity Distribution Analysis.}
We analyze the cosine similarity $\phi(p, c)$ between prediction and context across 7,371 test samples from ICEWS14 using the contrastively tuned model. The average similarity score for correct predictions exceeds that of incorrect ones by $\Delta\mu = 0.057$. This supports our assumption that similarity can serve as a proxy for semantic plausibility.

\paragraph{Novelty vs. Utility.}
We further observe that:
\begin{itemize}
    \item 9.1\% of predictions are non-historical despite the gold answer being present in $\mathcal{H}$.
    \item Among all non-historical predictions, only 1.5\% are correct and improve Hits@10.
\end{itemize}

These findings indicate that many model generations deviate from the historical context unnecessarily and fail to yield substantial gains. They motivate fallback to more salient entities when regeneration fails.

\paragraph{Threshold Optimization.}
The optimal threshold $\tau^*$ is learned by maximizing separation between correct ($\mathcal{C}$) and incorrect ($\mathcal{I}$) prediction similarities:
\begin{equation}
\tau^* = \arg\max_\tau \left[ F_\mathcal{C}(\tau) - F_\mathcal{I}(\tau) \right]
\end{equation}
where $F$ is the empirical CDF of cosine similarity values over samples from $\mathcal{C}$ and $\mathcal{I}$.

\paragraph{Fallback Scoring.}
If generation fails after $k$ iterations (we use $k=1$), the model selects a final answer from $\mathcal{H}$ using:
\begin{align}
f(h) &= \frac{\text{count}(h)}{|\mathcal{H}|}, \\
r(h) &= 1 - \frac{\text{pos}(h)}{|\mathcal{H}|}, \\
\psi(h) &= \beta \cdot f(h) + (1 - \beta) \cdot r(h),
\end{align}
where $\text{pos}(h)$ denotes the rank of $h$ in its occurrence order. We set $\beta = 0.6$ in all experiments.

We compute cosine similarities between predicted entities and prompt context using the \texttt{all-mpnet-base-v2} sentence transformer from HuggingFace. The threshold $\tau^*$ is tuned on a development set by maximizing the separation between correct and incorrect predictions.

\begin{figure}[t]
  \centering
  \includegraphics[trim=6 5 10 10, clip, width=\linewidth]{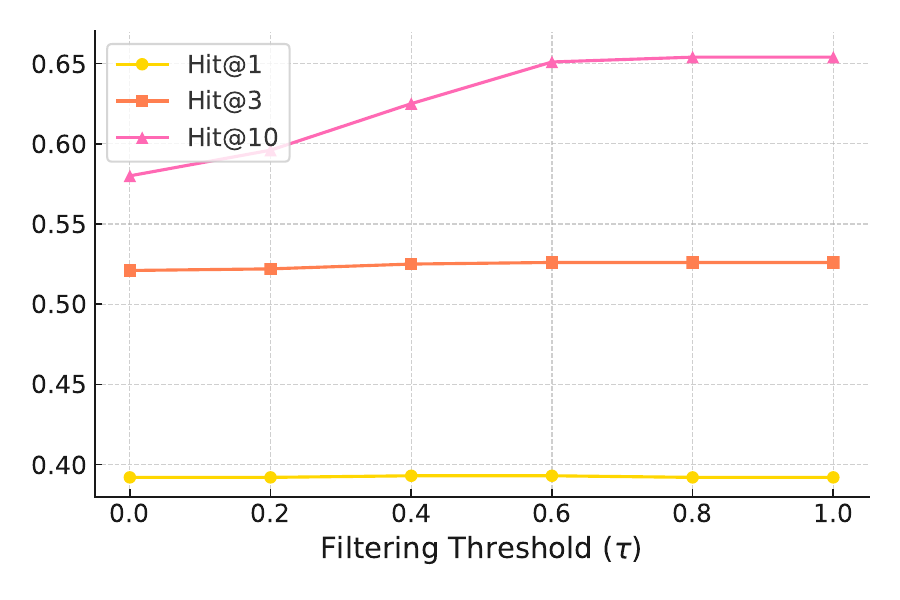}
  \caption{Effect of filtering threshold ($\tau$)}
  \label{fig:filtering_threshold}
\end{figure}

Figure~\ref{fig:filtering_threshold} examines the effect of the semantic filtering threshold $\tau$. As the threshold increases, Hits@10 improves, peaking near $\tau = 0.6$. Always falling back to historical entities ($\tau = 1.0$) slightly increases accuracy at the cost of exploration and computational efficiency. Threshold $\tau = 0.6$ balances correction with flexibility, enabling the model to revise low-quality outputs without overconstraining its generation space. 

\section{Baseline Model Details}
\label{appendix:baseline_details}

We compare RECIPE-TKG against several baseline methods that reflect the dominant modeling paradigms for TKG forecasting. Embedding-based methods include RE-GCN~\cite{li2021temporal}, which applies relational graph convolutions to timestamped graph snapshots; xERTE~\cite{han2020xerte}, which combines subgraph sampling and path-based reasoning using attention for explainability; TANGO~\cite{han2021learning}, which uses neural ODEs to learn continuous-time entity embeddings; and TimeTraveler~\cite{sun2021timetraveler}, which employs reinforcement learning to explore multi-hop temporal paths. Rule-based method includes TLogic~\cite{liu2022tlogic} relies on extracted symbolic rules for forecasting. The results of these models are derived from ~\citet{liao2024gentkg}

We also replicate two recent LLM-based methods. ICL~\cite{lee2023tkgicl} applies in-context learning by prepending historical quadruples to a query and using greedy decoding with a regex-based answer extraction. GenTKG~\cite{liao2024gentkg} performs parameter-efficient fine-tuning with LoRA adapters, and combines this with a rule-based history sampling module. We use their official codebases and replicate their evaluation pipelines for fair comparison.

\section{Dataset Statistics}
\label{appendix:dataset_stats}

We use four standard temporal knowledge graph benchmarks. ICEWS14 and ICEWS18 are subsets of the Integrated Crisis Early Warning System, containing geopolitical event records with daily granularity. GDELT provides global political event data, filtered to the most frequent events for tractability. YAGO consists of curated facts from a multi-year period. The statistics for these datasets are provided in Table~\ref{tab:dataset_stats}.

\begin{table*}[ht]
\centering
\renewcommand{\arraystretch}{1.1}
\setlength{\tabcolsep}{5pt}
\caption{Dataset statistics used in our experiments. Time granularity varies by dataset and influences temporal resolution.}
\begin{tabular}{l|ccc|cc|c}
\toprule
\rowcolor{SuperSoftPink}
\textbf{Dataset} & \textbf{\#Train} & \textbf{\#Valid} & \textbf{\#Test} & \textbf{\#Entities} & \textbf{\#Relations} & \textbf{Time Gap} \\
\midrule
ICEWS14 & 74{,}845 & 8{,}514 & 7{,}371 & 7{,}128 & 230 & 1 day \\
ICEWS18 & 373{,}018 & 45{,}995 & 49{,}545 & 23{,}033 & 256 & 1 day \\
GDELT   & 79{,}319 & 9{,}957 & 9{,}715 & 5{,}850 & 238 & 15 mins \\
YAGO    & 220{,}393 & 28{,}948 & 22{,}765 & 10{,}778 & 24 & 1 year \\
\bottomrule
\end{tabular}
\label{tab:dataset_stats}
\end{table*}

\section{More Analysis}
\subsection{Analysis of Contrastive Fine-Tuning}
\label{appendix:contrastive_analysis}
To complement the ablation results in Section~\ref{sec:ablation}, we analyze how contrastive fine-tuning affects model behavior in low-history regimes—settings where standard exact-match metrics such as Hits@k may fail to capture the semantic relevance of model predictions.

\paragraph{Setup.} We group ICEWS14 test samples by history length and compute the semantic distance between each model prediction and the gold entity. We compare three supervision settings: ICL, SFT, and contrastive FT, all evaluated under the same TLR history sampling.

We define semantic distance using cosine similarity between predicted and gold entities in a sentence embedding space:
\begin{equation}
\phi(p, o) = 1 - \text{cos-sim}(E(p), E(o)),
\end{equation}
where $E(\cdot)$ denotes the sentence transformer used in Section~\ref{sec:filtering}. Lower $\phi$ indicates higher semantic alignment, even if the prediction does not exactly match the gold entity.

\begin{figure*}[h]
\centering
\includegraphics[width=0.83\linewidth]{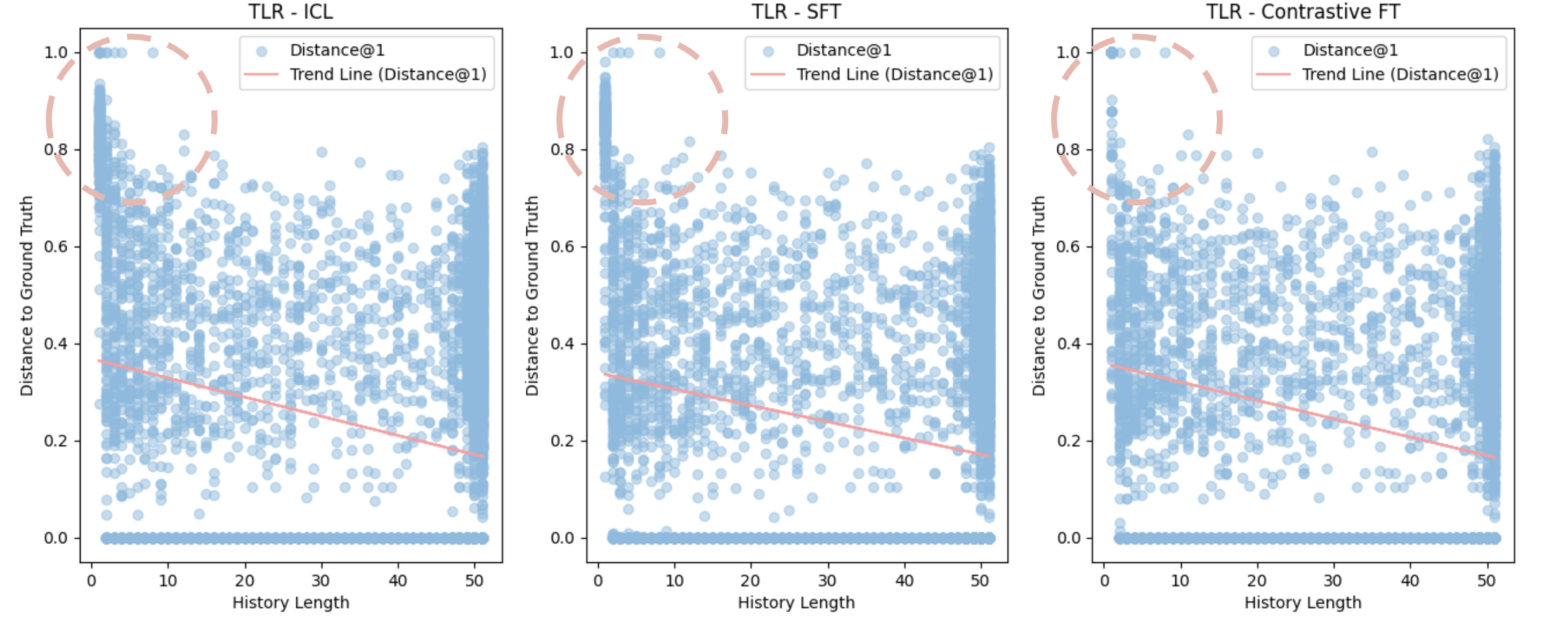}
\caption{Semantic distance ($\phi$) vs. history length on ICEWS14 under TLR sampling. The encircled region highlights CL’s improved semantic grounding in sparse-history settings.}
\label{fig:history_vs_distance}
\end{figure*}

\paragraph{Contrastive Tuning Improves Semantic Grounding.}
Figure~\ref{fig:history_vs_distance} plots the semantic distance $\phi(p, o)$ against the retrieved history length. All models show the expected trend: greater history generally yields predictions closer to the gold entity in embedding space. However, the distinction between supervision strategies becomes clear in low-history regimes. In the encircled region (history length $\leq 3$), contrastive fine-tuning produces fewer high-distance predictions than both ICL and SFT. This demonstrates that contrastive learning enhances the model’s ability to infer plausible entities even when the input lacks strong historical evidence.

\paragraph{Multi-hop Sampling Further Stabilizes Model Behavior.}
To examine how our sampling strategy affects model reasoning on sparse-history inputs, we repeat the same experiment using our proposed \textit{RBMH Sampling}. For comparability, we compute semantic distances on the same subset of samples originally identified as short-history under TLR.

\begin{figure*}[h]
\centering
\includegraphics[width=\linewidth]{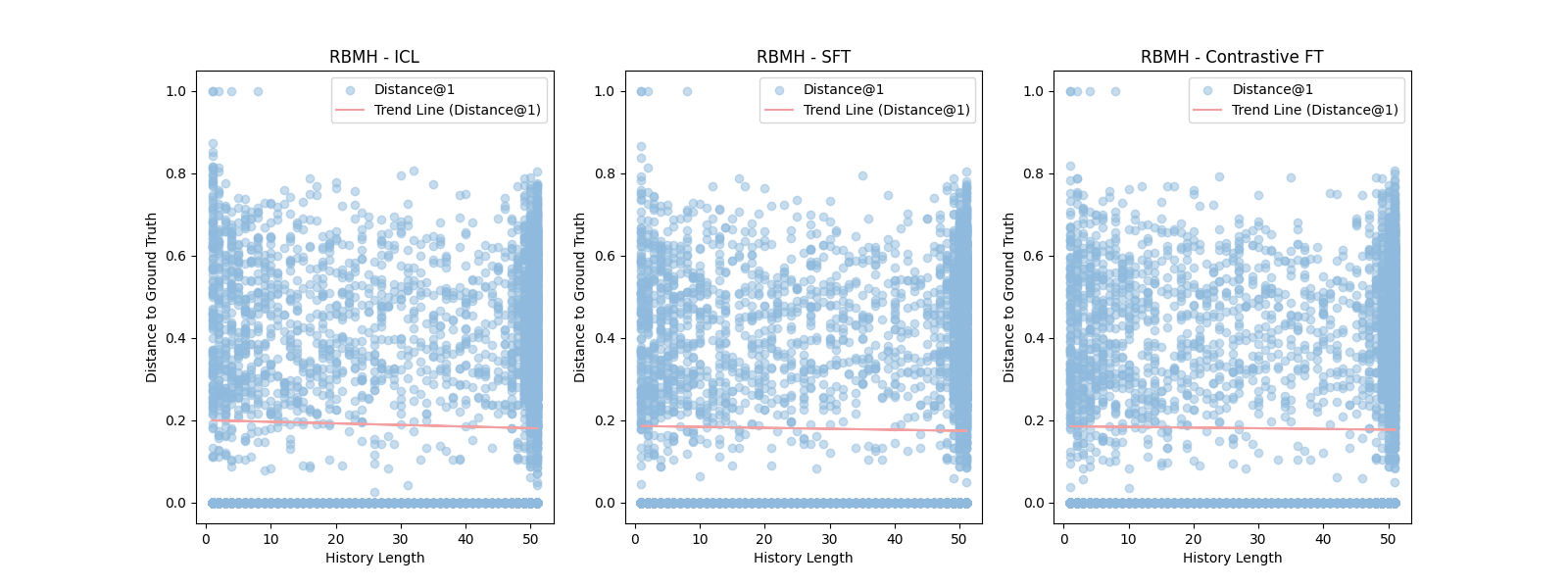}
\caption{Semantic distance ($\phi$) vs. history length for the same TLR-identified sparse samples, but evaluated under \textit{RBMH Sampling}. The model exhibits more stable behavior across history lengths.}
\label{fig:history_vs_distance_rbmh}
\end{figure*}

As shown in Figure~\ref{fig:history_vs_distance_rbmh}, contrastive-tuned models under \textit{RBMH Sampling} exhibit more uniform semantic behavior across history lengths. Unlike the steep drop-off observed under TLR, the semantic distance remains relatively stable, indicating that many samples previously limited by shallow context can now be grounded through richer structural and temporal cues. This supports our motivation in Section~\ref{challenges:grounding}: one-hop sampling often fails to provide the necessary relational evidence, and multi-hop expansion is essential for enabling reliable reasoning, rather than the test instances being inherently harder.

\begin{figure*}[h]
\centering
\includegraphics[width=\linewidth]{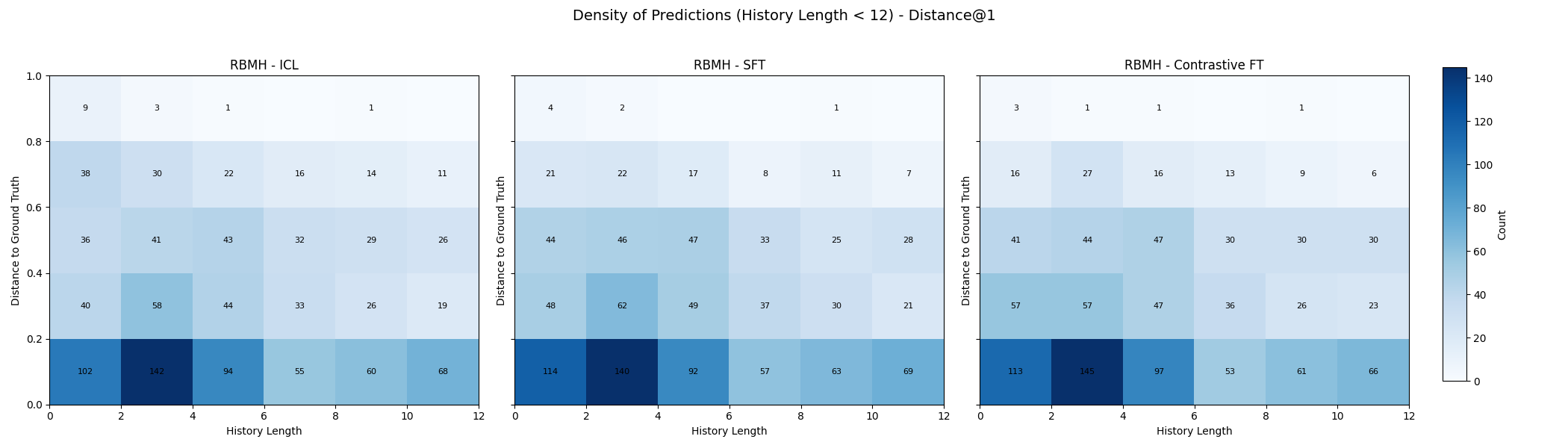}
\caption{Semantic distance ($\phi$) vs. history length for the same TLR-identified sparse samples, but evaluated under \textit{RBMH Sampling}. \textit{CFT} learns better with RBMH as it samples the deeper relationships between entities.}
\label{fig:cl_rbmh}
\end{figure*}

\paragraph{Qualitative Support.}
Figure~\ref{fig:cl_rbmh} presents qualitative examples where contrastive-tuned models produce predictions that are not exact matches but remain relationally and contextually appropriate. In contrast, ICL and SFT often produce surface-level or unrelated completions. These examples, paired with the distributional evidence above, underscore how contrastive fine-tuning improves semantic generalization and interpretability, particularly when Hits@k offers limited signal.

\begin{comment}

\begin{table}[h]
\centering
\renewcommand{\arraystretch}{1.1}
\caption{Sample predictions on ICEWS14 with no retrieved history. CL yields more plausible outputs.}
\resizebox{\linewidth}{!}{
\begin{tabular}{l|p{4.5cm}|p{3.5cm}}
\toprule
\rowcolor{SuperSoftPink}
\textbf{Model} & \textbf{Prediction} & \textbf{Gold Entity} \
\midrule
ICL & \texttt{UN Peacekeepers} & \texttt{Military_Personnel_USA} \
SFT & \texttt{Foreign_Ministry_France} & \texttt{Military_Personnel_USA} \
CL & \texttt{US Defense Officials} & \texttt{Military_Personnel_USA} \
\bottomrule
\end{tabular}
}
\label{tab:contrastive_examples}
\end{table}
\end{comment}

\paragraph{Case Study.}
To better understand the behavior of RECIPE-TKG, we provide a case study comparing the top-10 predictions of four methods on a specific query. The ground-truth object is \texttt{High\_Ranking\_Military\_Personnel\_(Nigeria)}, which is not explicitly present in the history. As shown in Figure~\ref{fig:case_study}, none of the models are able to perfectly predict the correct entity. However, the predictions made by RECIPE-TKG models are clearly more semantically aligned with the ground truth. For example, predictions such as \texttt{Military\_(Nigeria)} and \texttt{Defense\_Personnel\_(Nigeria)} closely approximate the true answer in meaning, whereas other models (ICL and GenTKG) fail to capture such relevant semantics. This demonstrates the advantage of contrastive fine-tuning in shaping the embedding space, allowing the model to produce more relationally compatible predictions even when exact matches are not observed in history.

\begin{figure*}[t]
\centering
\begin{blackbox}[Model Outputs]
\footnotesize
\begin{minipage}[t]{0.495\textwidth}
\textbf{ICL-\texttt{LLaMA2-7b}} \\
1. \texttt{Citizen\_(Nigeria)} \\
2. \texttt{Boko\_Haram} \\
3. \texttt{Suleiman\_Abba} \\
4. \texttt{Other\_Authorities\_/\_Officials\_(Nigeria)} \\
5. \texttt{Aliyu\_Mohammed\_Gusau} \\
6. \texttt{Nigerian\_Army} \\
7. \texttt{Nigerian\_Army} \\
8. \texttt{Nigerian\_Army} \\
9. \texttt{Nigerian\_Army} \\
10. \texttt{Other\_Authorities\_/\_Officials\_(Nigeria)} \\
[2pt]
\hdashrule{\linewidth}{0.5pt}{2pt 2pt} \\[3pt]
\textbf{RECIPE-TKG-\texttt{LLaMA2-7b}} \\
1. \texttt{Citizen\_(Nigeria)} \\
2. \texttt{Boko\_Haram} \\
3. \texttt{Suleiman\_Abba} \\
4. \texttt{Other\_Authorities\_/\_Officials\_(Nigeria)} \\
5. \texttt{Aliyu\_Mohammed\_Gusau} \\
6. \texttt{Government\_(Nigeria)} \\
7. \texttt{Military\_(Nigeria)} \\
8. \texttt{Abdul\_Aziz\_Yari} \\
9. \texttt{Chief\_of\_Staff\_(Nigeria)} \\
10. \texttt{Abdul\_Aziz\_Yari} \\
\end{minipage}
\hfill
\begin{minipage}[t]{0.48\textwidth}
\textbf{GenTKG-\texttt{LLaMA2-7b}} \\
1. \texttt{Citizen\_(Nigeria)} \\
2. \texttt{Boko\_Haram} \\
3. \texttt{Suleiman\_Abba} \\
4. \texttt{Other\_Authorities\_/\_Officials\_(Nigeria)} \\
5. \texttt{Nigeria} \\
6. \texttt{Aliyu\_Mohammed\_Gusau} \\
7. \texttt{Nigeria} \\
8. \texttt{Nigeria} \\
9. \texttt{Nigeria\_Army} \\
10. \texttt{None} \\
[2pt]
\hdashrule{\linewidth}{0.5pt}{2pt 2pt} \\[3pt]
\textbf{RECIPE-TKG-\texttt{LLaMA3-8b}} \\
1. \texttt{Citizen\_(Nigeria)} \\
2. \texttt{Other\_Authorities\_/\_Officials\_(Nigeria)} \\
3. \texttt{Boko\_Haram} \\
4. \texttt{Suleiman\_Abba} \\
5. \texttt{Defense\_/\_Security\_Ministry\_(Nigeria)} \\
6. \texttt{Terrorist\_(Boko\_Haram)} \\
7. \texttt{Employee\_(Nigeria)} \\
8. \texttt{Terrorist\_(Nigeria)} \\
9. \texttt{Senior\_Military\_Official\_(Nigeria)} \\
10. \texttt{Defense\_Personnel\_(Nigeria)} \\
\end{minipage}

\vspace{6pt}
\textbf{Ground-truth entity:} \texttt{High\_Ranking\_Military\_Personnel\_(Nigeria)}
\end{blackbox}
\caption{Top-10 predictions from four models. RECIPE-TKG produce semantically closer outputs to the ground truth.}
\label{fig:case_study}
\end{figure*}

\section{Use of AI Tools}
AI assistants were used to support writing (e.g., phrasing suggestions) and code generation (e.g., syntax templates). All such outputs were subject to thorough human verification, and the authors remain fully responsible for the content presented.

\end{document}